\documentclass[10pt,twocolumn,letterpaper]{article}

\usepackage{cvpr}              

\newcommand{\cmark}{{\color{OliveGreen} \ding{51}}}
\newcommand{\xmark}{{\color{BrickRed} \ding{55}}}

\definecolor{Gray}{gray}{0.9}
\definecolor{Red}{RGB}{255, 184, 179}
\definecolor{Green}{RGB}{202, 235, 213}

\newcommand{\cpar}[1]{%
  \begingroup
    \ifthenelse{\lengthtest{#1 pt > 0pt}}{%
      \textcolor{black}{(}\textcolor{OliveGreen}{+#1}\textcolor{black}{)}%
    }{%
      \ifthenelse{\lengthtest{#1 pt = 0pt}}{%
        \textcolor{black}{(}\textcolor{gray}{+0.00}\textcolor{black}{)}%
      }{%
        \textcolor{black}{(}\textcolor{BrickRed}{#1}\textcolor{black}{)}%
      }%
    }%
  \endgroup
}

\newcommand{\cparper}[1]{%
  \begingroup
    \ifthenelse{\lengthtest{#1 pt > 0pt}}{%
      \textcolor{black}{(}\textcolor{OliveGreen}{+#1\%}\textcolor{black}{)}%
    }{%
      \ifthenelse{\lengthtest{#1 pt = 0pt}}{%
        \textcolor{black}{(}\textcolor{gray}{+0\%}\textcolor{black}{)}%
      }{%
        \textcolor{black}{(}\textcolor{red}{#1\%}\textcolor{black}{)}%
      }%
    }%
  \endgroup
}

\usepackage{multirow}
\usepackage{listings}
\usepackage{bbm}
\usepackage{amsmath}
\usepackage{amssymb,algorithm}
\usepackage{algpseudocode}
\usepackage{amssymb}
\usepackage{pifont}
\usepackage{xcolor}
\usepackage{ifthen}
\usepackage{cancel}
\usepackage[normalem]{ulem}
\usepackage{colortbl}
\usepackage{rotating}

\definecolor{cvprblue}{rgb}{0.21,0.49,0.74}
\usepackage[pagebackref,breaklinks,colorlinks,allcolors=cvprblue]{hyperref}

\def\paperID{35292} 
\def\confName{CVPR}
\def\confYear{2026}

\title{RealBirdID: Benchmarking Bird Species Identification in the Era of MLLMs}

\author{Logan Lawrence \quad \quad Mustafa Chasmai \quad \quad Rangel Daroya \quad \quad Wuao Liu \quad \quad Seoyun Jeong \\
Aaron Sun \quad Max Hamilton \quad Fabien Delattre \quad Oindrila Saha \quad Subhransu Maji \quad Grant Van Horn \\[.5ex]
Computer Vision Lab, UMass Amherst\\
{\tt\small \{lclawrence, mchasmai, rdaroya, wuaoliu, seoyunjeong,} \\
{\tt\small aaronsun, jmhamilton, fdelattre, osaha, smaji, gvanhorn\}@umass.edu}\\[1ex] \textbf{\href{https://github.com/cvl-umass/RealBirdID}{\texttt{github.com/cvl-umass/RealBirdID}}}}

\begin{document}
\maketitle

\begin{abstract}

    Fine-grained bird species identification in the wild is frequently unanswerable from a single image: key cues may be non-visual (e.g. vocalization), or obscured due to occlusion, camera angle, or low resolution. Yet today’s multimodal systems are typically judged on answerable, in-schema cases, encouraging confident guesses rather than principled abstention. We propose the RealBirdID benchmark: given an image of a bird, a system should either answer with a species or abstain with a concrete, evidence-based rationale: ``requires vocalization," ``low quality image," or ``view obstructed". For each genus, the dataset includes a validation split composed of curated unanswerable examples with labeled rationales, paired with a companion set of clearly answerable instances. We find that (1) the species identification on the answerable set is challenging for a variety of open-source and proprietary models ($\leq$$13\%$ accuracy for MLLMs including GPT-5 and Gemini-2.5 Pro), (2) models with greater classification ability are not necessarily more calibrated to abstain from unanswerable examples, and (3) that MLLMs generally fail at providing correct reasons even when they do abstain. RealBirdID establishes a focused target for abstention-aware fine-grained recognition and a recipe for measuring progress.
    
\end{abstract}
\vspace{-.5cm}
\begin{figure}[t!]
    \centering
    \captionsetup{type=figure}
    \setlength{\belowcaptionskip}{-5pt}
    \includegraphics[width=\linewidth]{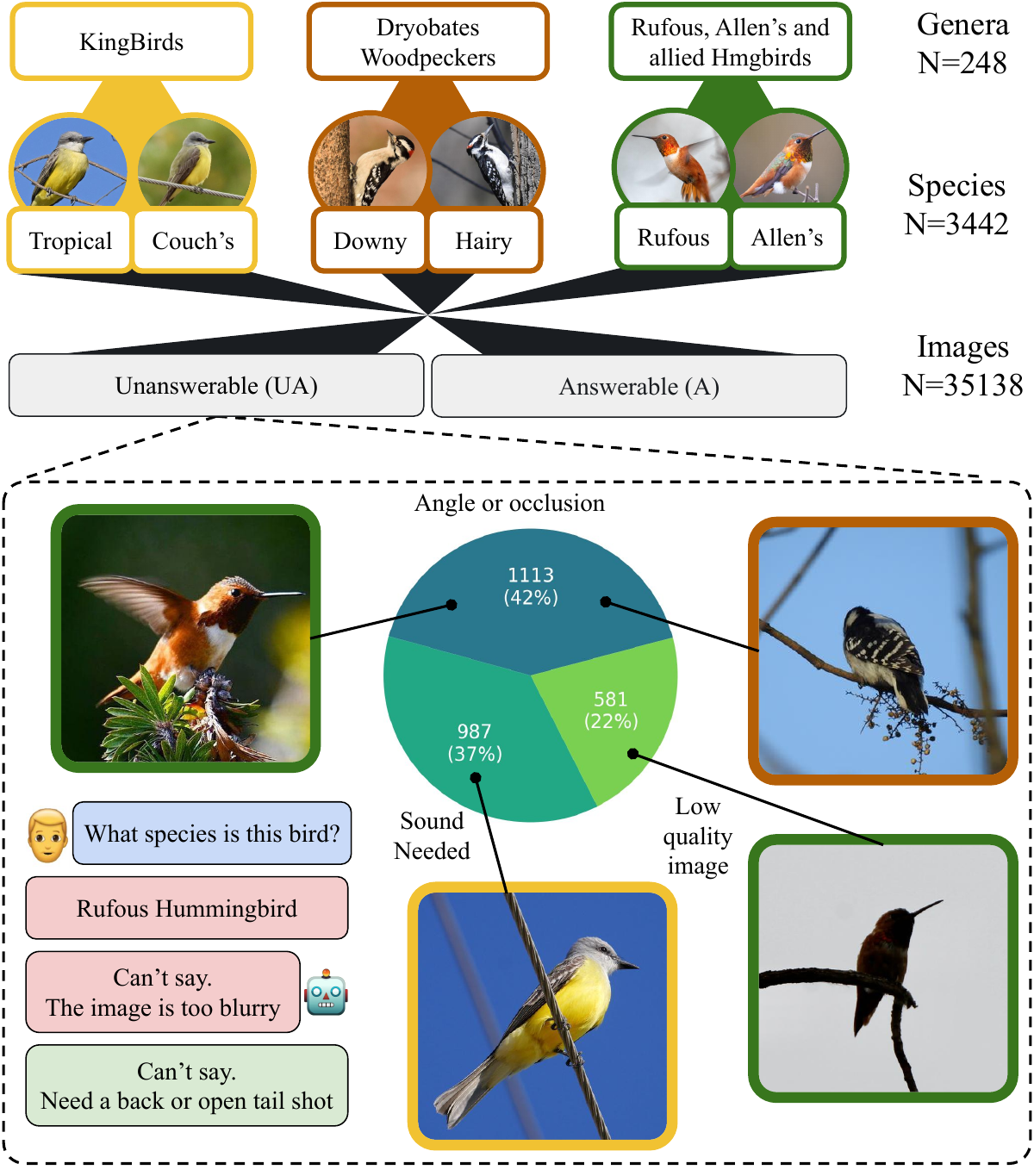}
    \vspace{-20pt}
    \captionof{figure}{\textbf{Preview of RealBirdID.} In contrast to previous species identification datasets, in each of the genus associated with RealBirdID there is \textbf{a corresponding set of unanswerable (UA) examples}. The summary metric proposed gauges both (1) the ability for the classifier to disambiguate between confusing classes and (2) abstain from predicting on unanswerable examples. Incorrect abstention reasoning is shown in \colorbox{Red}{red} whereas correct reasoning is shown in \colorbox{Green}{green} .
    }
    \label{fig:paper_preview}
\vspace{-.3cm}
\end{figure}

\section{Introduction}

\begin{figure*}[ht!]
    \centering
    \captionsetup{type=figure}
    \setlength{\belowcaptionskip}{-10pt}
    \includegraphics[width=\linewidth]{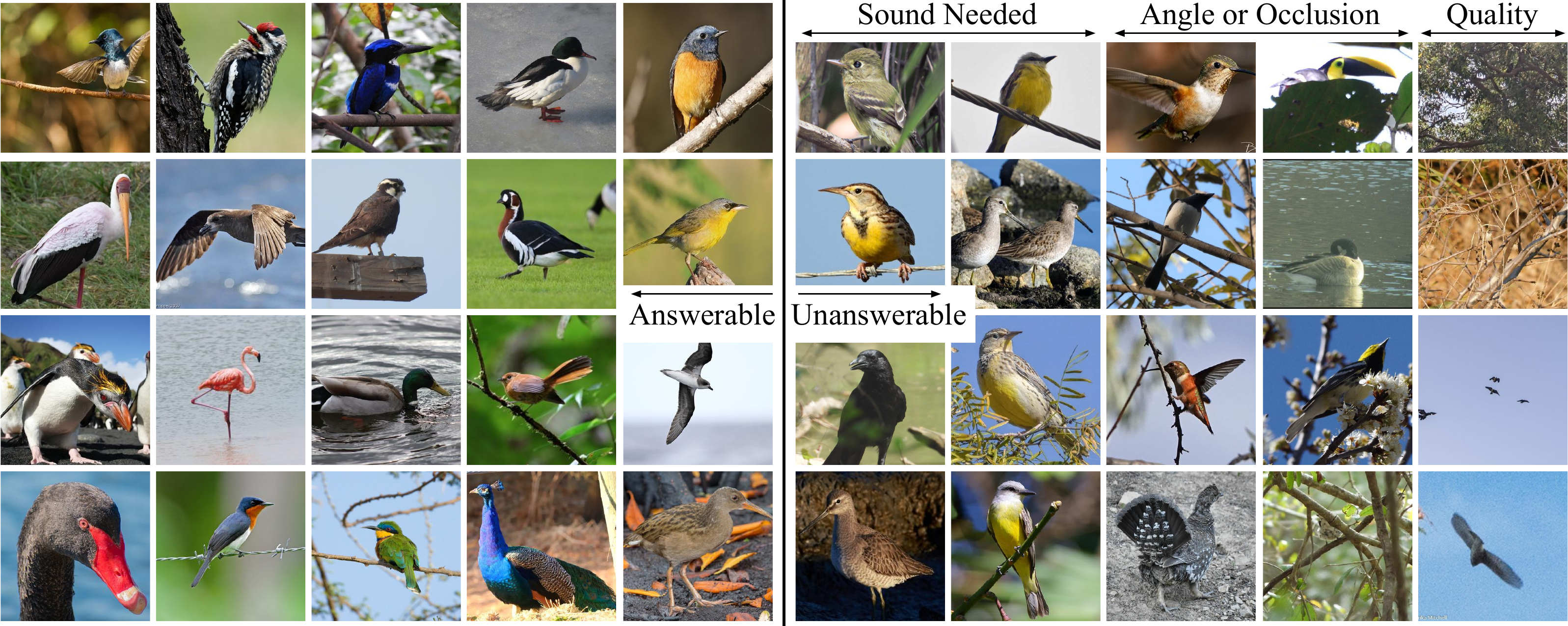}
    \vspace{-20pt}
    \captionof{figure}{\textbf{Peek into the dataset}.  Few examples for the answerable and unanswerable images in RealBirdID. Unanswerable examples are grouped by possible unanswerable reasons. 
    Images from ``Sound needed" may be harder to abstain than ``Quality" for example, as more detailed knowledge about the particular genus would be needed to understand these reasons. 
    }
    \label{fig:preview_combined}
\end{figure*}

Bird species identification has been crucial to measuring progress in fine-grained visual recognition (FGVR), with benchmarking capabilities advancing via datasets like CUB-200-2011 \cite{cub_dataset}, NABirds \cite{nabirds_dataset}, NeWT \cite{newt_dataset}, INQUIRE \cite{inquire_dataset}, and the iNaturalist challenges \cite{inat_dataset} which emphasize parts, attributes, and expert taxonomic structure. In particular, zero-shot visual classification has been revitalized by the introduction of Multimodal Large Language Models (MLLMs) \cite{gpt4v, instructblip, llava, llava_next}. These developments, along with open-vocabulary prompting and instruction-tuned multimodal supervision have pushed FGVR performance upward across fully supervised, few-shot, and zero-shot settings \cite{hong2025unlabeled, he2025analyzing, saha2024improved, saha2025generate, lawrence2025you}, enabling systems to handle increasingly subtle intra-class distinctions.

However, real world deployment scenarios are often out-of-distribution or, even worse, out-of-schema from existing benchmarks \cite{bendale2015towards, koh2021wilds, bendale2016towards}. In particular, MLLMs are often trained with datasets comprised exclusively of answerable examples, whereas it is common to see examples in deployment that cannot be properly classified \cite{vqav2_dataset, rgqa_dataset}. Being forced to choose an answer can cause hallucinations \cite{kalai2025language, huang2025survey} which can be misleading and potentially dangerous, especially in settings like medicine or law \cite{thirunavukarasu2023large, guha2023legalbench, wu2023bloomberggpt, tomani2024uncertainty}. 


In such cases, one would like the model to give a reason for \textit{abstaining} from picking one of the predefined classes, which could give the opportunity for human experts to intervene and refine the system. The idea of abstention has recently gained traction within text-only tasks \cite{kirichenko2025abstentionbench, tomani2024uncertainty, madhusudhan2024llms}. However, the study of abstention of MLLMs in visual settings is still in its infancy \cite{rgqa_dataset}. To the best of our knowledge, there has been no work focused on measuring abstention \textit{when keeping text fixed}. Namely, the model can only \textit{visually reason} in order to abstain.


To address this gap, we introduce the \textbf{RealBirdID} dataset, a benchmark for evaluating the ability of MLLMs to exhaustively predict fine-grained species for a given genus, as well as the ability to abstain under \textit{realistic, unanswerable examples} for that genus (\cref{fig:paper_preview}). We come to the following conclusions:
\vspace{.15cm}

\noindent
\textbf{1. Exhaustive species identification is an unsolved problem in vision-language systems.} When thresholding on model uncertainty, models generally do not exceed $0.3$ Area Under the Curve (AUC) on species-level prediction on the answerable set, corresponding to roughly $17\%$ accuracy (\cref{table:combined_auc_q1}). Using range map information to restrict the list of species under consideration significantly increases classification performance across all models (\cref{fig:range_map_effect}).

\vspace{.15cm}
\noindent
\textbf{2. Current methods have difficulty forming tradeoffs between classification and abstention.} Furthermore, they are barely better than random at separating unanswerable and answerable examples (\cref{tab:summary_new}). Using range maps only has mild effects on abstention tradeoff (\cref{fig:range_map_effect}). In fact, using range info for MLLMs actually \textit{degrades} abstention performance. 

\vspace{.15cm}
\noindent
\textbf{3. Even when models abstain, their reasons are often incorrect.} Across all models, we observed a strong bias in abstention reasons (\cref{table:abstention_confusion_recall}), in some cases foregoing vocalization completely (\cref{fig:abstention_confusion}). MLLMs abstention rates are very brittle in the face of wording variation on instructions (\cref{fig:brittle_mllm}), sometimes ranging from 3\% to 23\% abstention rate under semantically equivalent prompt wordings.

\vspace{.15cm}

\section{Related Work} 

\paragraph{Benchmarking Abstention Capabilities} In text-only tasks, the idea of \textit{unanswerability} has gained traction (e.g. SQuAD2.0 \cite{squad2.0}, NewsQA \cite{newsqa}, BigBench (KnownUnknowns) \cite{bigbench}, KUQ \cite{amayuelas2023knowledge}, VQAv2 \cite{vqav2_dataset}). However, these methods do not consider the \textit{reason} for being unable to answer when given an example, which we refer to as \textit{abstention with reasoning,} and will be focusing on in this work. 



AbstentionBench \cite{kirichenko2025abstentionbench} introduces a 35k-example benchmark spanning 20 datasets to evaluate whether LLMs appropriately abstain on unanswerable, underspecified, subjective, or outdated questions, and shows that scaling models offers little benefit. SelfAware \cite{yin2023large} benchmarks LLM self-knowledge by contrasting unanswerable and answerable questions with an automated uncertainty-detection protocol, finding models can flag ``unknowns" above chance yet remain overconfident with category and model-dependent variance. In contrast to the \textit{text-only} datasets AbstentionBench and SelfAware, \textit{we focus on inputs with images} as containing the evidence for abstention.

RGQA \cite{rgqa_dataset} establishes a realistic VQA (RVQA) benchmark by pairing standard answerable questions with 29k human-annotated unanswerable questions (fine- and coarse-grained) to test rejection and answering jointly. In contrast to RGQA, we specifically focus on \textit{needing visual information to abstain rather than detecting unanswerable texts}. Namely, the prompts in our data are identical or semantically equivalent (some meaning akin to \emph{``What is the species of bird in this image?"}), and the models evaluated are forced to reason about visual content to know to abstain. 

Finally, \citet{Snaebjarnarson_2025_CVPR} propose hierarchical classification metrics which revealed how current vision-language models misalign predictions across hierarchy levels. Similarly,  ~\citet{tan2025vision} show that Vision-LLMs systematically underperform in hierarchical reasoning through their proposed metric, attributing this limitation to their language backbone rather than the visual encoder. However, both of these works assume that each node in the taxonomy has a ground truth label. In our problem setting, the taxonomic tree is \textit{partially-labeled}, eluding most metrics.


\vspace{.2cm}
\noindent
\textbf{Hierarchical Classification and Classifier Abstention} Predicting or evaluating within a taxonomic hierarchy is an emerging direction similar to abstention, where models predict across a label scheme with an implicit tree structure. \cite{wang2023transhp, park2024visually, novack2023chils, alper2024emergent, xia2025hgclip, ren2023chatgpt}. Contrary to most works in this area, we wish to measure how well vision encoders are able to correctly predict the genus even when the species may be unanswerable, namely gauge the ability of the classifier \textit{to stop prediction in intermediate nodes in the taxonomy} rather than always at the leaf nodes. 

With growing interest in explainable AI, a number of methods explore better capture of uncertainty in  prediction scores~\cite{hu2023uncertainty, cole2023selectively, kuhn2023semantic, szarvas2012cross, farkas2010conll} or verbal articulation~\cite{xiao2022uncertainty, zhou2023navigating, lin2022teaching, yin2024reasoning}. A tangential line of research explores post-hoc abstention in models not specifically designed for interpretability. 
Elliciting confidences from LLMs free-form responses has shown strong calibration with favorable risk–coverage tradeoffs \cite{kadavath2022language, jiang2021can, jiang2020can, sharifi2022should}, but these works focus on text-only tasks. Besides the ability to capture uncertainty, the decision making process for abstention is not well understood. Explicit reasons for unanswerability in RealBirdID allow us to assess the model's alignment with human abstention.

\begin{figure}[t!]
    \centering
    \captionsetup{type=figure}
    \setlength{\belowcaptionskip}{-10pt}
    \includegraphics[width=\linewidth]{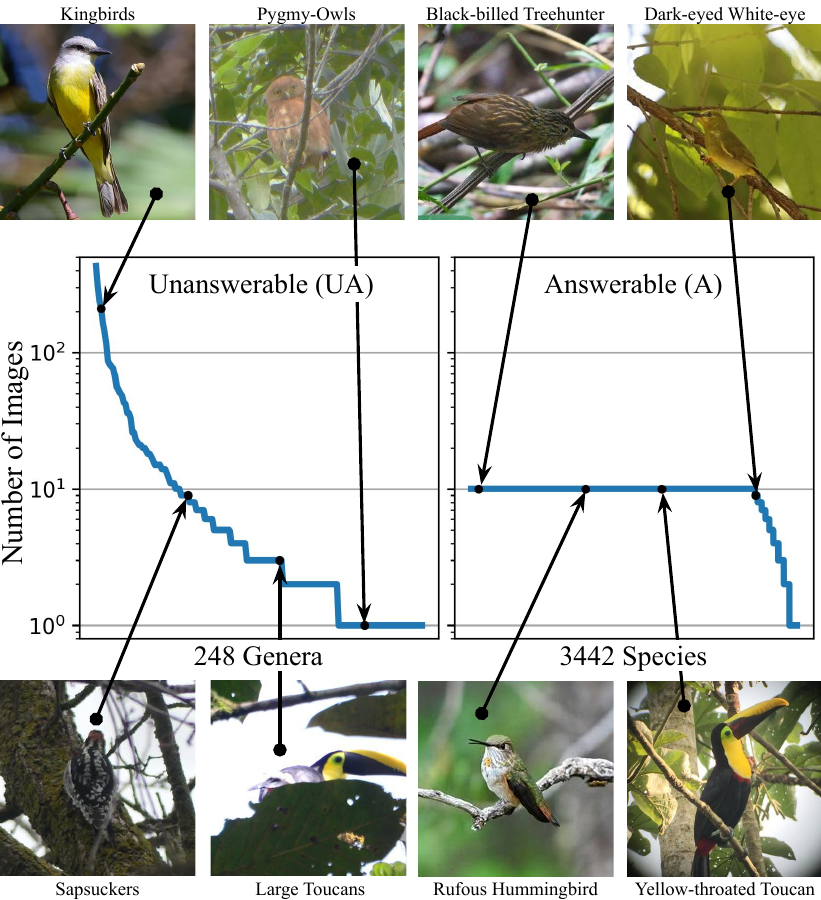}
    \vspace{-20pt}
    \caption{\textbf{Distribution of images across the answerable (A) and unanswerable (UA) subsets in RealBirdID.} The genera within the unanswerable (UA) subset exhibit a highly imbalanced, long-tailed distribution. For example, 85\% of UA images originate from just 61 genera that each contain more than five unanswerable samples. For details on the distribution of the 3,442 species, see \cref{appendix:sec:remaining_figures}.}
    \label{fig:longtailed_species}
\end{figure}


\section{The RealBirdID Benchmark}

\begin{figure*}[t]
\centering
\hspace{-225px}
\begin{subfigure}[t]{.55\textwidth}
  \centering
  \includegraphics[width=\linewidth]{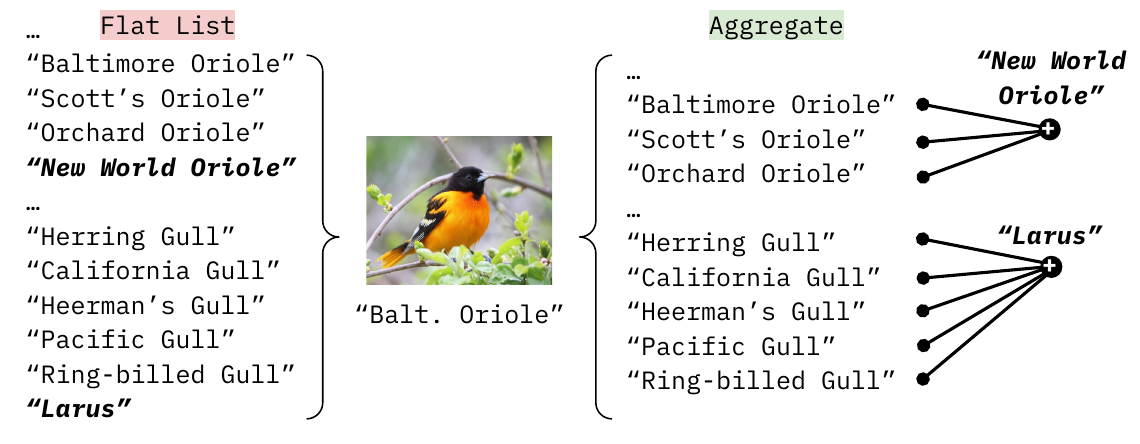}
  \label{fig:sub1}
\end{subfigure}%
\hspace{100px}
\begin{subtable}[h!]{.70\textwidth}
\vspace{-135px}
\setlength{\tabcolsep}{9pt}
\hspace{205px}
{\scriptsize
    \begin{tabular}{lccc}
    \toprule
     \textbf{Method} & \textbf{Answerable} & \textbf{Unanswerable} & \textbf{HM} \\
     \\[-2ex]
    \hline
    \rowcolor{Gray} &&& \\[-1.8ex]
    \rowcolor{Gray}
    \multicolumn{4}{l}{CLIP-L/14} \\[.3ex]
     \\[-2ex]
    Flat List	& 0.0	&7.1	&0.0 \\
    TreeGT & \textbf{6.7 \cpar{6.7}}	& \textbf{11.9 \cpar{4.7}}	& \textbf{8.5 \cpar{8.5}} \\
     \\[-2ex]
    \hline
    \rowcolor{Gray} &&& \\[-1.8ex]
    \rowcolor{Gray}
    \multicolumn{4}{l}{MetaCLIP-L/14} \\[.3ex]
     \\[-2ex]
    Flat List	& 0.1	&5.3	&0.2\\
    TreeGT	& \textbf{9.7 \cpar{9.6}}	&\textbf{16.1 \cpar{10.8}}	& \textbf{12.1 \cpar{11.9}} \\
        \\[-2ex]
    \hline
    \rowcolor{Gray} &&& \\[-1.8ex]
    \rowcolor{Gray}
    \multicolumn{4}{l}{SigLIP-SO400m/14} \\[.3ex]
     \\[-2ex]
    Flat List	&0.0	&7.1	&0.0 \\
    TreeGT	& \textbf{6.7 \cpar{6.7}}	& \textbf{11.9 \cpar{4.7}}	& \textbf{8.5 \cpar{8.5}} \\
    \bottomrule
    \end{tabular}
}
\end{subtable}
\vspace{-20px}
\caption{\textbf{Multiple choice question (MCQ) formatting is a problem for encoder models.} A straightforward implementation of an abstention class with CLIP models is to simply expose the genus level as a text prompt and treat its prediction as abstention. However, this approach greatly underperforms a modification from a previous hierarchical method, TreeGT \cite{deng2012hedging}. ``\textbf{HM}" refers to the harmonic mean of the ``\textbf{Answerable}" and ``\textbf{Unanswerable}" accuracies. Bird photographed by Tom Murray on iNaturalist (\href{https://www.inaturalist.org/photos/155708903}{link}).}
\label{fig:mcq_problems_v2}
\end{figure*}

\subsection{Collecting Unanswerable (UA) Examples}


\textbf{iNaturalist}\footnote{\url{https://www.inaturalist.org/}} is a global biodiversity citizen-science platform where participants upload geo-tagged observations of wild organisms that are collaboratively identified by the community, producing open data widely used in ecology and conservation. An observation first becomes \textit{verifiable} if it passes initial quality checks (e.g., having a date, location, supporting media). A subset of verifiable observations are then promoted to \textit{Research Grade} when the community reaches at least a two-thirds consensus on an identification at species level or lower or when the community agrees that the maximum taxonomic depth has been reached.

\vspace{.2cm}
\noindent
\textbf{Initial Sampling} We queried the iNaturalist API for \emph{verifiable} bird observations without temporal or geographic restriction (see \cref{appendix:sec:remaining_figures} for observed date distributions). We intentionally did \emph{not require} Research Grade since our goal is to find observations with genuine disagreement at the species level. This stands in contrast to previous datasets using iNaturalist which are comprised of exclusively Research Grade images, mostly having identifications at the species level or below. For summary of how RealBirdID contrasts with previous datasets, see \cref{table:rbid_comparison}.

Starting from 1.4M candidate observations at the \textit{verifiable} level, we retained the first image per observation and applied a two-stage prefilter: (i) bird presence via YOLOv3~\cite{redmon2018yolov3} and (ii) perceptual quality using MANIQA~\cite{yang2022maniqa}. This produced 410k images containing a bird and passing the quality gate. Next, we filtered observations to those which had at least one genus-level prediction and two contributors. Finally, we filtered out any observations which contained annotations indicating the image was of a dead bird, egg, or feather. We also note as a limitation that some multi-image observations feature a clearer bird only in later frames. We leave multi-frame handling for future work.

\vspace{.2cm}
\noindent
\textbf{Finding Unanswerable Examples}
For the remaining observations, we extracted the full comment and identification history and parsed ambiguity signals with a library of hand-crafted regular expressions and lightweight heuristics. Each matched pattern maps to a provisional schema of \emph{abstention reasons} (e.g., distributional/range ambiguity, life-stage or sex dimorphism, molt/wear, insufficient viewpoint, hybrid/escapee, taxonomic uncertainty, and low image quality). We retained observations whose community taxon resolved to at most the genus level and for which at least one valid reason was detected, yielding 5{,}300 ``unanswerable'' exemplars. For an example of some of these images, see \cref{fig:paper_preview} and \cref{fig:preview_combined}.

\vspace{.2cm}
\noindent
\textbf{Quality Control}
A subset of the parsed data was verified by an expert annotator equipped with Birds of the World\footnote{\url{https://birdsoftheworld.org}} Field Identification, a popular field tool for disambiguating similar bird species. The annotator was asked to verify previously extracted ``Vocalization," ``Angle / Occlusion," and ``Image Quality" failures. The annotator also discarded observations if they were too difficult to decide on or if the discussion or identifications contained inaccuracies. This created a subset of labeled data which was then used to further refine text parsing. This iterative process yielded $3.4$k examples with parsed abstention reasons.




\subsection{Collecting Answerable (A) Examples}\label{subsec:collecting_answerable_examples}

\paragraph{Exhaustive Species Sampling}
For each unanswerable observation with a community taxon above species (typically genus), we enumerated all descendant species using the iNaturalist Taxon API and sampled \emph{Research Grade} images for each species. We targeted up to 200 images per species (hard cap), without additional balancing at this stage. This procedure induces the expected long-tailed distribution across species (see \cref{fig:longtailed_species}). Because this release serves as a validation resource rather than a training benchmark, we did not create train/val/test splits.

\vspace{.2cm}
\noindent
\textbf{Likely Species Determination}
To surface the most plausible candidates under contention for each unanswerable case, we derive a geo-contextual species checklist using SINR \cite{cole2023spatial}, which takes in spatial coordinates $\mathbf{x} = [lon, lat]$ and produces a probability vector of species occurring at that location. Data which did not have location information available is kept but marked as being ineligible for species list constriction. A total of $3148$ unanswerable and $25773$ answerable observations were parsed for likely species.



\begin{table}
\caption{\textbf{RealBirdID vs. related fine-grained bird benchmarks.} ``iNat19-Birds" refers to the validation split of iNat19 subset to the Aves class (namely only birds). ``\textbf{\#A}" and ``\textbf{\#UA}" refer to the number of \textit{answerable} and \textit{unanswerable} images, respectively. }
\vspace{-.2cm}
\centering
\footnotesize
\setlength{\tabcolsep}{8pt}
\begin{tabular}{l@{\hskip .2cm}cccc}
\textbf{Dataset} & \textbf{Has Train Set?} & \textbf{\# A} & \textbf{\# UA} & \textbf{\# Species} \\
\midrule
NABirds & \cmark & 24633 & 0 & 555 \\
CUB-200 & \cmark & 5794 & 0 & 200 \\
iNat19-Birds & \cmark & 14860 & 0 & 1486 \\
RealBirdID  & \xmark & 31885 & 3253 & 3442 \\
\bottomrule
\end{tabular}
\label{table:rbid_comparison}
\end{table}

\section{Abstention Metrics for Encoders}



In the classical zero-shot setting, encoder models (e.g., CLIP) do not have an abstention class. Hence, when operating in hierarchical settings there is no way to implicitly tell when a model wishes to predict a leaf node in a hierarchy or one of the intermediate nodes. To demonstrate this, we provide a baseline using the concatenated list of species and genera (\cref{fig:mcq_problems_v2}), where we find that the flat list method vastly underperforms a naive version of probability aggregation, TreeGT \cite{deng2012hedging}.

To overcome this, many methods derive an abstention criterion from the predicted distribution over species. A prediction is reclassified as an abstention if the model's predictive uncertainty (e.g., top-1 probability, top-1 margin, or entropy) fail to satisfy the specified thresholds. In our experiments, we use the top-1 class probability as the abstention criterion. We experiment with different choices of abstention criterion in \cref{appendix:sec:metric_details}. One can use softmax probabilities over species to calculate higher ranking taxa probabilities by summing from leaf to parent nodes. For example, the probability of the ``Crows and Ravens" genus would be the sum of the probabilities from the $53$ species occurring within that genus. However, this process only gives abstention decisions when applying a \textit{specific} threshold. Instead, we propose to evaluate models by \textit{sweeping this threshold} and aggregating the tradeoffs into a summary metric.

\begin{figure}[t!]
    \centering
    \captionsetup{type=figure}
    \setlength{\belowcaptionskip}{-15pt}
    \includegraphics[width=\linewidth]{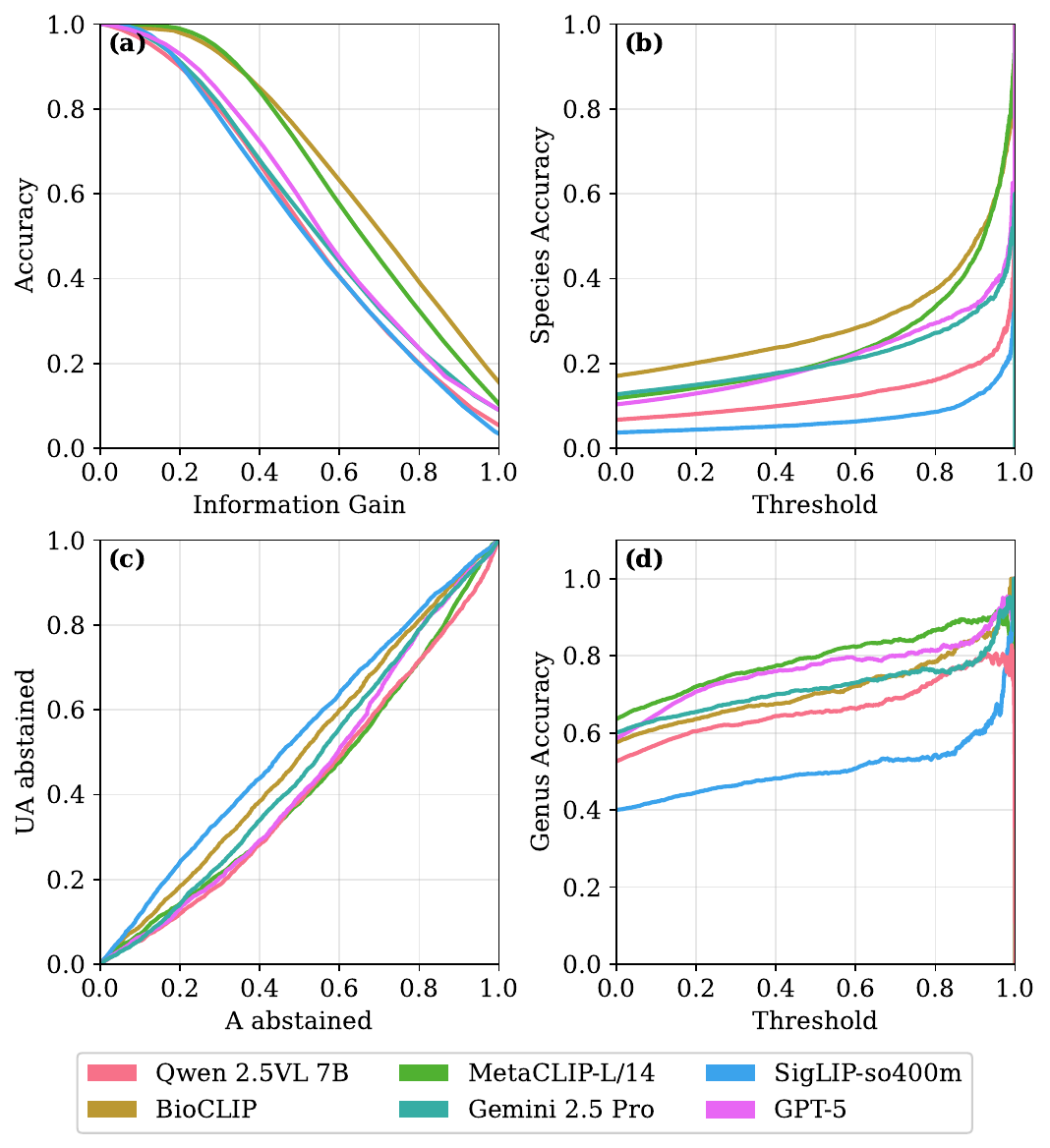}
    \vspace{-15pt}
    \captionof{figure}{\textbf{Visualization of sweeping parameters for classification and abstention metrics on popular CLIP-based models and MLLMs.} To summarize how deep in the hierarchical classifiers can go while staying accurate we use Information Gain vs. Accuracy \textbf{(a)}. Each classifier admits a tradeoff curve when predicting species \textbf{(b)} and genus \textbf{(d)}. To measure separation between the unanswerable and answerable set, we measure the AUC of the model entropy \textbf{(c)}.}
    \label{fig:visualization_metrics_new}
\end{figure}

\vspace{.2cm}
\noindent
\textbf{Metric 1: Abstention Tradeoff (UA/A).} We consider the fraction of images that a model abstains on with a particular threshold for both answerable and unanswerable samples. By changing this threshold, one forms a tradeoff curve between abstention on known answerable and unanswerable instances. An ideal model would abstain for all unanswerables and not abstain for any answerable. For an illustrative example, we point to \cref{fig:visualization_metrics_new}(c): curves which are closer to the upper-left indicate better separation of UA and A. We refer to the area under these curves as the \textit{abstention tradeoff} or \emph{UA/A}.






\vspace{.2cm}
\noindent
\textbf{Metric 2: Classification Performance (IG).}\label{sec:performance_metrics} In order to avoid rewarding models which trivially abstain for every output, we need a metric which takes into account the classification accuracy of the model. However, our data suffers from partial unanswerability: some of the images simply \textit{cannot be assigned to species} within the hierarchy. To solve this, we make use of the Accuracy vs. Information Gain curves as proposed by DARTS \cite{deng2012hedging}, which considers predictions at all levels of the hierarchy. 
Given a probability vector across species, we aggregate probabilities to compute predictions at the genus level, class level, \etc. 
By sweeping across a range of threshold values, we evaluate each prediction based on two criteria: its accuracy (correct vs incorrect) and its information gain, which corresponds to the taxonomic depth of the prediction (e.g., species vs genus). Plotting these results yields a tradeoff curve, illustrating how different thresholds can increase predictive accuracy at the cost of reduced taxonomic precision, e.g. \cref{fig:visualization_metrics_new}(a).
The area under these curves (referred to as \textit{classification performance} or \emph{IG}) summarizes this tradeoff. 


\vspace{.2cm}
\noindent
\textbf{Metric 3: Model Calibration (AUC).} We propose an additional metric for measuring model calibration. For a given abstention fraction, we calculate the accuracy for answerable images. As an illustrative example, we point to \cref{fig:visualization_metrics_new}(b)(d), where we show accuracy at various thresholds on max probability. For the answerable set, we look at the species and genus accuracies while for the unanswerable set, we look at the genus accuracy. These plots are summarized by their areas (AUC) and presented in Table~\ref{table:combined_auc_q1}  alongside species and genus accuracies (Acc) for the full dataset. We use micro-averaged accuracy here, but also report macro-averaged (over species or genera) and class-imbalance weighting in \cref{appendix:sec:additional_experiments}.6. Together, these summarize how accuracy trades off with abstention without fixing an operating point.



\section{Experiments}
\begin{table*}[]
    \centering
    \setlength{\tabcolsep}{4pt}
    \caption{\textbf{Summary of classification and abstention performance on RealBirdID.} ``IG" and ``UA/A" refer to the proposed Metrics 1 and 2 in \cref{sec:performance_metrics}. InternVL3, Qwen-2.5VL, Gemma3, and Llama-3.2V are their 8B, 7B, 12B, and 11B instruction-tuned variants. Best performance for each metric is \textbf{bolded} whereas the next highest is \underline{underlined}.}
    \vspace{.3cm}
    {\footnotesize
        \begin{tabular}{lccccccccc@{\hskip -1px}c@{\hskip 10px}c@{\hskip -2px}c@{\hskip -2px}c@{\hskip -2px}c@{\hskip -2px}c@{}c}
          & \multicolumn{3}{c}{\textbf{CLIP}} & \multicolumn{3}{c}{\textbf{MetaCLIP}} & \multicolumn{2}{c}{\textbf{WildCLIP}} & \\
        \cmidrule(lr){2-4}  
        \cmidrule(lr){5-7}
        \cmidrule(lr){8-9}
         \textbf{Metric} \hspace{10px} & \rotatebox{0}{\textbf{B/32}} & \rotatebox{0}{\textbf{B/16}} & \rotatebox{0}{\textbf{L/14}} & \rotatebox{0}{\textbf{B/32}} & \rotatebox{0}{\textbf{B/16}} & \rotatebox{0}{\textbf{L/14}} & \textbf{Base} & \textbf{Lite} & \multirow[t]{2}{*}{\rotatebox{45}{\textbf{SigLIP}}} & \multirow[t]{2}{*}{\rotatebox{45}{\textbf{BioCLIP}}} & \multirow[t]{2}{*}{\rotatebox{45}{\textbf{InternVL3}}}& \multirow[t]{2}{*}{\rotatebox{45}{\textbf{Qwen-2.5VL}}} & \multirow[t]{2}{*}{\rotatebox{45}{\textbf{Gemma-3}}} & \multirow[t]{2}{*}{\rotatebox{45}{\textbf{Llama-3.2V}}} & \multirow[t]{2}{*}{\rotatebox{45}{\textbf{Gemini-2.5P}}} & \multirow[t]{2}{*}{\rotatebox{45}{\textbf{GPT-5}}} \\
        \midrule     
        IG & 50.6 & 55.3 & 62.0 & 54.2 & 59.1 & \underline{66.0} & 55.1 & 53.3 & 53.7 & \textbf{68.9} & 46.5 & 54.2 & 46.3 & 48.4 & 56.4 & 57.7 \\
        UA/A & 47.8 & 44.6 & 48.1 & 47.2 & 44.7 & 42.5 & 46.7 & \underline{48.4} & 53.2 & \textbf{49.6} & 45.6 & 41.7 & 39.2 & 43.0 & 46.2 & 44.1 \\
        \bottomrule
        \end{tabular}
    }
    \label{tab:summary_new}
    \vspace{-7px}
\end{table*}

\vspace{.2cm}
\noindent
\textbf{Models Considered} For encoder models, we consider BioCLIP \cite{stevens2024bioclip},  WildCLIP \cite{gabeff2024wildclip}, CLIP \cite{radford2021learning}, MetaCLIP \cite{xu2023demystifying}, and SigLIP \cite{zhai2023siglip}. We choose these due to their public availability, popularity, as well as lack of overlap with the construction of RealBirdID. For example, we find that the training set of BioCLIP-2 \cite{gu2025bioclip}, TreeOfLife200M, overlaps with $57\%$ of the answerable set, thus we do not consider it. For further details, see \cref{appendix:sec:additional_experiments}. We use the same base prompt, \texttt{"a photo of a \{species\} bird."} across the full-species list ($3651$ prototypes), then aggregate to higher rank taxa by summing across child nodes. This forms a hierarchy where each level has normalized probabilities, making it amenable for the proposed metrics.

\begin{table}[t]
\setlength{\belowcaptionskip}{-25pt}
\setlength{\tabcolsep}{4.5pt}
\begin{center}
\vspace{.95cm}
\captionof{table}{\textbf{Summary of classification performance of popular vision-language models in terms of taxonomic classification metrics.} Columns report Area Under Curve (visualization in \cref{fig:visualization_metrics_new}) and classification accuracy at the Species and Genus levels for the Answerable set and at the Genus level for the Unanswerable set. MLLMs are their instruction-tuned variant.}
{\footnotesize
\begin{tabular}{lcccccc}
 & \multicolumn{4}{c}{\textbf{Answerable}} & \multicolumn{2}{c}{\textbf{Unanswerable}}\\
    \cmidrule(lr){2-5}
    \cmidrule(lr){6-7}
    &&& &&& \\[-2ex]
 & \multicolumn{2}{c}{Species} & \multicolumn{2}{c}{Genus} & \multicolumn{2}{c}{Genus}\\
 & AUC & Acc  & AUC & Acc & AUC & Acc\\
&&& &&&  \\[-2ex]
\hline
\rowcolor{Gray} \hspace{2.1cm} &&& &&&\\[-1.8ex]
\rowcolor{Gray}
\textbf{Encoder-Based} &&& &&& \\[.3ex]
  &&& &&& \\ [-2ex]
CLIP-B/32 & 09.9 & 04.6 & 46.8 & 33.8 & 55.2 & 40.3 \\
CLIP-B/16  & 12.3 & 05.8 & 53.3 & 40.3 & 63.2 & 46.5 \\
CLIP-L/14 & 15.5 & 07.7 & 60.7 & 48.7 & 65.9 & 54.0 \\
MetaCLIP-B/32 & 15.4 & 06.8 & 55.4 & 39.0 & 64.0 & 46.4 \\
MetaCLIP-B/16 & 18.9 & 08.5 & 61.4 & 45.7 & 71.6 & 54.9 \\
MetaCLIP-L/14 & 25.1 & 11.8 & \textbf{70.4} & 56.1 & \textbf{79.0} & \textbf{63.6} \\
WildCLIP & 08.2 & 04.3 & 38.1 & 32.1 & 43.5 & 37.3 \\
WildCLIP$_{lite}$ & 08.7 & 04.1 & 43.7 & 33.2 & 48.9 & 37.1 \\
SigLIP-so400m & 07.2 & 03.7 & 49.0 & 36.7 & 50.7 & 40.0 \\
BioCLIP & \textbf{30.1} & \textbf{17.0} & 69.4 & \textbf{57.0} & 71.4 & 57.6 \\
&&& &&&  \\[-2ex]
\hline
\rowcolor{Gray} \hspace{2.1cm} &&& &&& \\[-1.8ex]
\rowcolor{Gray}
\textbf{Multimodal LLMs} &&& &&&  \\[.3ex]
&&& &&& \\[-2ex]
InternVL3-8B & 03.2 & 01.5 & 22.1 & 16.7 & 43.2 & 34.7 \\
Qwen2.5-VL-7B & 12.7 & 06.7 & 52.5 & 40.6 & 66.2 & 52.6 \\
Gemma-3-12B & 05.8 & 03.1 & 32.7 & 24.7 & 50.0 & 37.4 \\
Llama-3.2-11B-Vision & 09.5 & 04.6 & 39.4 & 28.8 & 55.4 & 42.7 \\
Gemini 2.5 pro & 21.4 & \textbf{12.7} & 66.9 & \textbf{52.8} & 71.8 & \textbf{60.1} \\
GPT-5 & \textbf{21.5} & 10.4 & \textbf{67.1} & 45.6 & \textbf{76.5} & 58.6 \\
\bottomrule
\end{tabular}
\label{table:combined_auc_q1}
}
\end{center}
\vspace{-20px}
\end{table}


For MLLMs, we consider Qwen2.5VL \cite{bai2025qwen25}, Llama-3.2V \cite{grattafiori2024llama}, Intern3-VL \cite{zhu2025internvl3}, PaliGemma \cite{beyer2024paligemma, steiner2024paligemma2}, Gemini-2.5 Pro \cite{comanici2025gemini25}, and GPT-5 \cite{gpt5}. For all MLLMs, unless otherwise stated, we use \texttt{"What is the species of this bird?"} as the prompt for obtaining a free-form response. We additionally use default parameters and use vLLM\footnote{\url{https://docs.vllm.ai/en/latest/}} to increase throughput. Each open-source model was run in a heterogeneous environment of single-GPU nodes, where each node required a GPU with at least 23GB VRAM, 16 CPU cores, and 64GB RAM. For the proprietary MLLMs GPT-5 and Gemini-2.5 Pro, we use the single-turn API calls using the same prompt \texttt{"What is the species of this bird?"} and the input image.




\vspace{.1cm}
\noindent
\textbf{Implementation Details} 
MLLM abstention detection and reasoning extraction was performed using nlg2choice \cite{lawrence2025you}, which first generates a free-form response to the question then extracts a final answer using constrained decoding. Specifically, for the second stage constrained decoding, we ask \texttt{"Does this text mention any of the following reasons: \{reason\_list\}"}. We additionally use the retrieval setting of nlg2choice to generate species probability vectors, which we aggregate up to the genus level using the leaf-parent node method mentioned previously. 




\section{Results}

\subsection{Zero-Shot Classification and Abstention}


\paragraph{Most encoder-based models struggle to perform well on classification.} We observe that, outside of the BioCLIP models, unthresholded accuracies of popular encoder models range from $3.7$-$17.0$\% at species level ($3442$-way) and $32.1$-$57.0$\% at genus level ($248$-way), depicted in \cref{table:combined_auc_q1}. Aligning with expectations, we find that model entropy correlates with difficulty in classification: accuracy increases as high entropy examples are dropped from the data. These curves yield calibration metrics consistent with this trend, depicted in \cref{fig:visualization_metrics_new}(b)(d).

\vspace{5px}
\noindent
\textbf{All models struggle to make tradeoffs between classification and abstention.} Abstention is a core requirement for reliable fine-grained recognition as models may achieve strong accuracy at both species and genus levels yet still be unable to abstain from prediction. In \cref{tab:summary_new}, we show the performance of various models on classification performance and ability to separate the UA and A set. We find that BioCLIP, in particular, has the highest classification performance $(IG = 68.9)$ but still lags behind SigLIP $(\textrm{UA/A} = 53.2)$ in terms of abstention tradeoff.


Overall, for the encoder models, we find no significant correlation between classification accuracy and abstention ability $(r_{pearson} = 0.60)$. Furthermore, we find that within the same training family, increasing model size and data corresponds to increases in classification performance $(\textrm{IG})$ but not to abstention tradeoff $(\textrm{UA/A})$. This suggests that abstention behavior is controlled by different factors than standard fine-grained recognition performance. Together, these results indicate that simply improving fine-grained accuracy on RealBirdID is insufficient for reliable abstention, and that encoder-based FGVR systems require explicit abstention-aware objectives or downstream selective-classification mechanisms to behave conservatively on unanswerable inputs.

\begin{figure}[t]
\centering
\vspace{-5px}
\begin{subfigure}[t]{.251\textwidth}
  \centering
  \includegraphics[width=\linewidth]{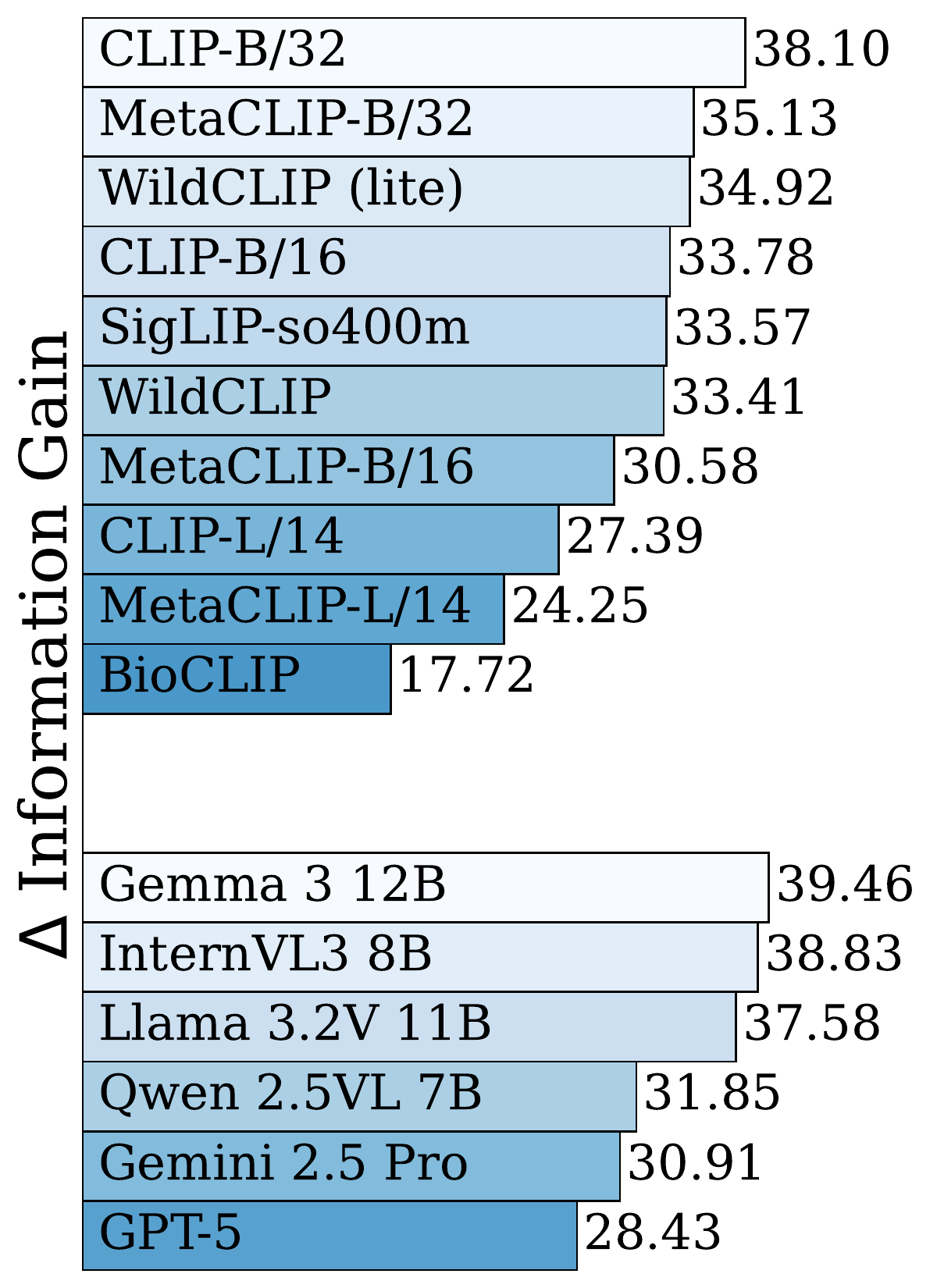}
  \label{fig:sub1}
\end{subfigure}
\begin{subfigure}[t]{.22\textwidth}
  \centering
  \includegraphics[width=\linewidth]{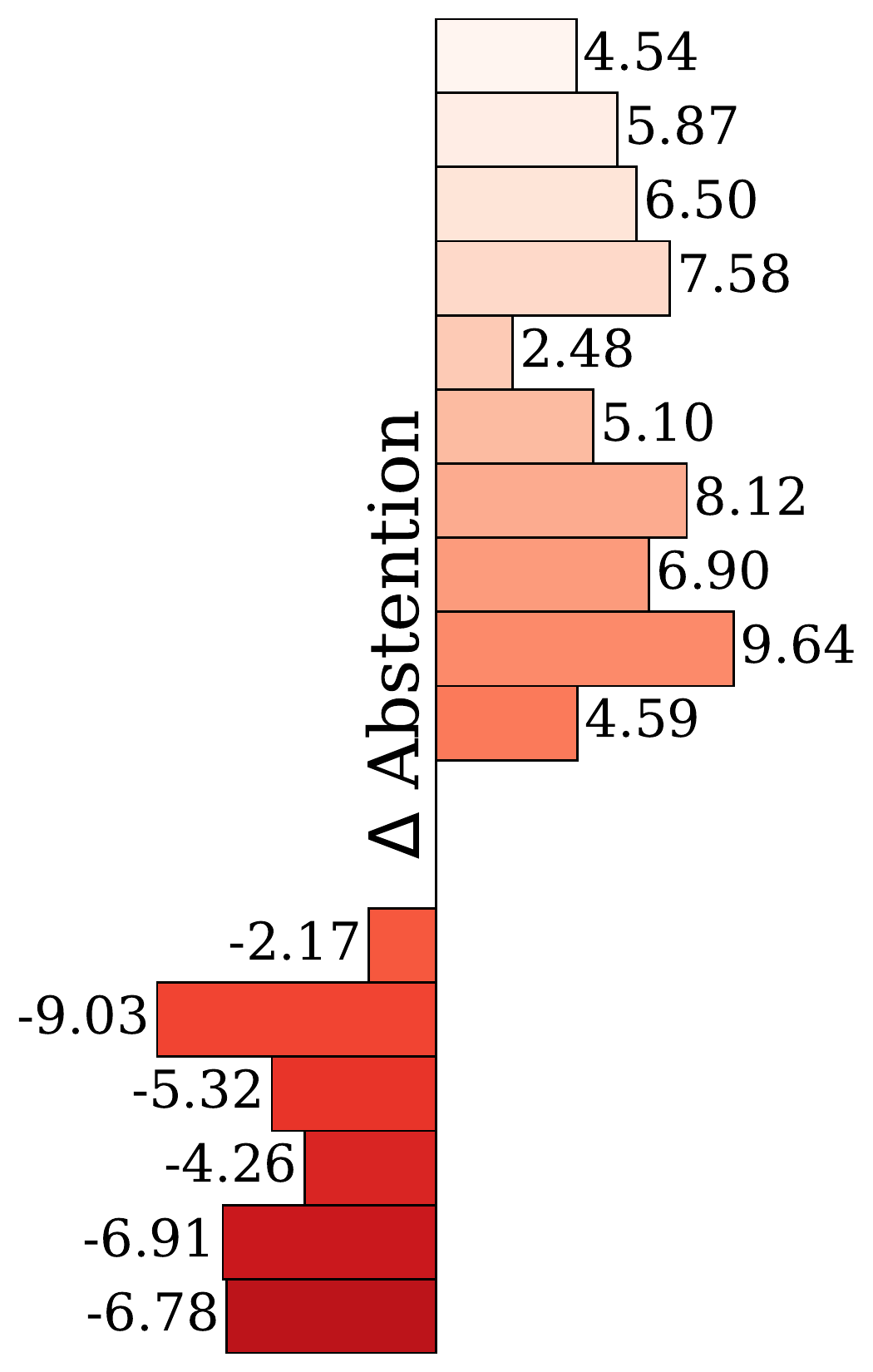}
  \label{fig:sub1}
\end{subfigure}
\vspace{-10px}
\caption{\textbf{Effect of range-map information on encoders and MLLMs.} Adding \textit{species range map} information increases classification ability for both encoders and MLLMs, as shown by positive information gain across all models. For encoders, it also increases abstention ability, namely by increasing separation between A / UA. Notably, abstention tradeoff performance decreases for MLLMs. We provide distributions and examples showing this in \cref{appendix:sec:additional_experiments}.}
\vspace{-15px}
\label{fig:range_map_effect}
\end{figure}
\begin{figure}[t]
    \centering
    \setlength{\belowcaptionskip}{-15pt}
    \captionsetup{type=figure}
    \includegraphics[width=\linewidth]{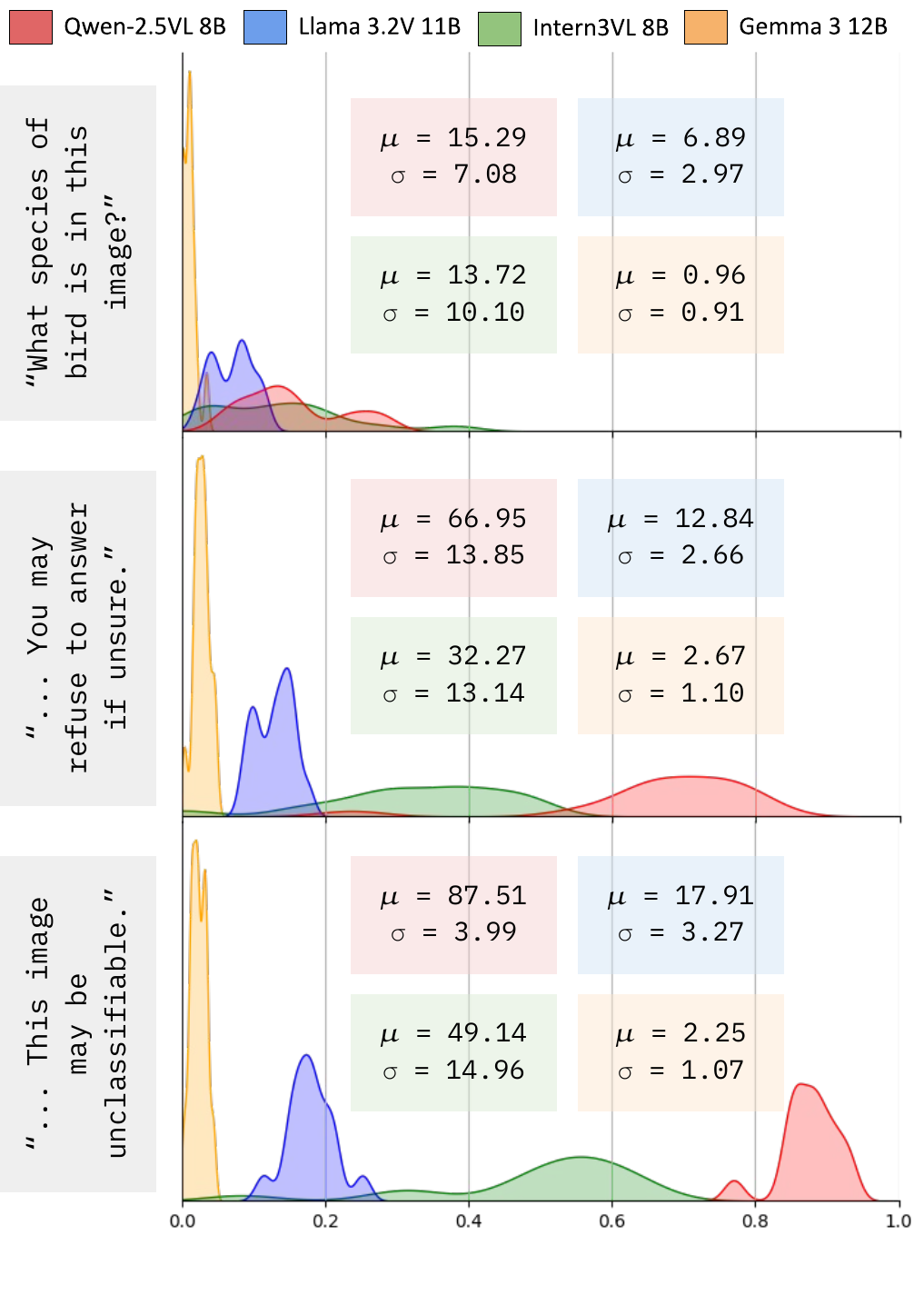}
    \vspace{-28pt}
    \captionof{figure}{\textbf{Abstention rate of MLLM on the UA set under equivalent writings of instructions.} Each distribution comprises 15 semantically equivalent ways of writing the same instruction, depicted in gray on the left.}
    \label{fig:brittle_mllm}
    \vspace{.1cm}
\end{figure}%
\begin{figure}[t]
    \centering
    \setlength{\belowcaptionskip}{-20pt}
    \captionsetup{type=figure}
    \includegraphics[width=\linewidth]{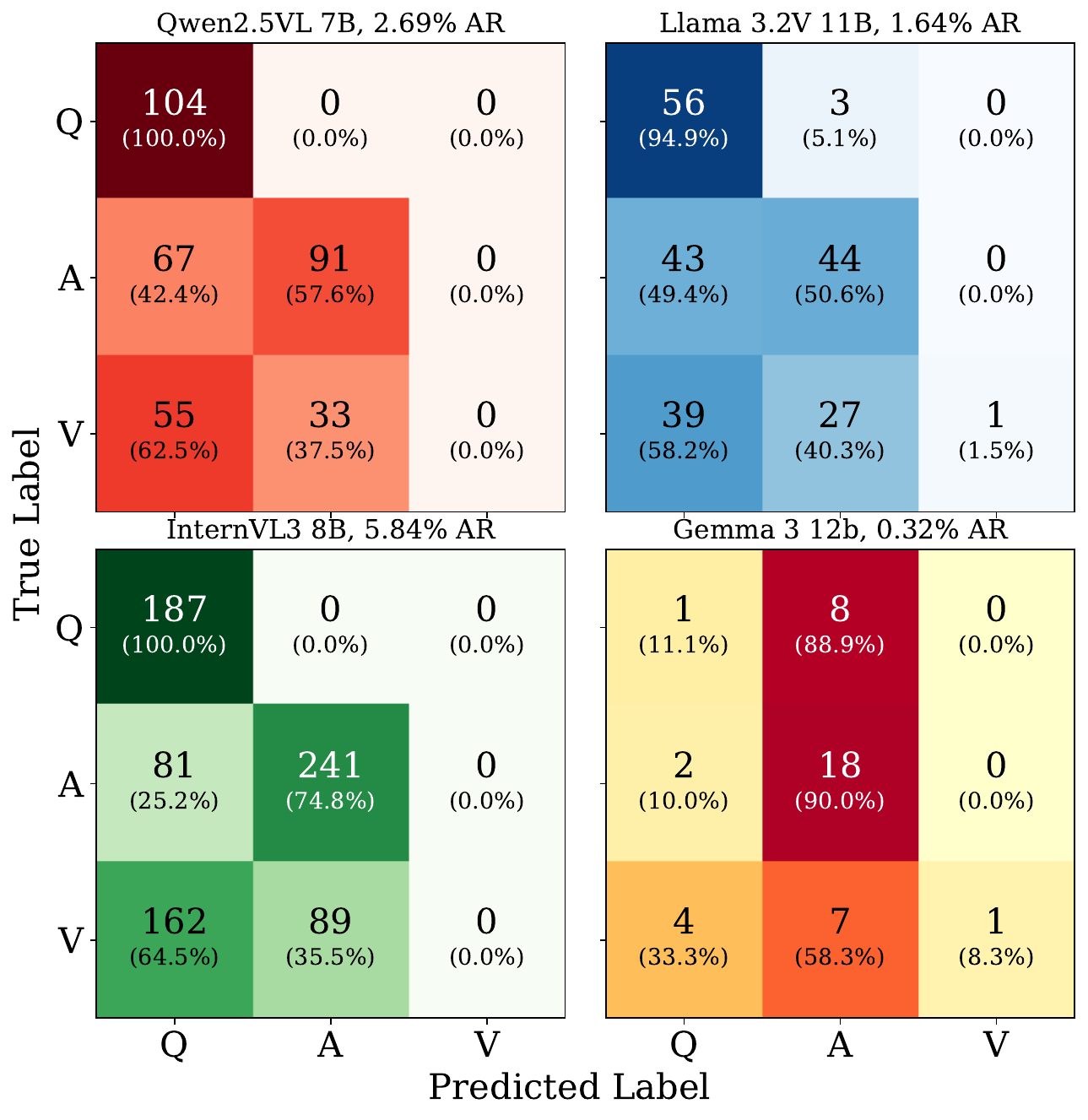}
    \vspace{-15pt}
    \captionof{figure}{\textbf{Confusion matrix of abstention reasons over various MLLMs.} For each MLLM, the confusion matrix is computed only over abstained examples, comparing the human-labeled cause: image quality issues (Q), angle or occlusion (A), or dependence on vocalization (V). Cells show counts with row-normalized percentages in parentheses, and each panel title reports the abstention rate (AR), the fraction of all validation examples on which the model abstains.}
    \label{fig:abstention_confusion}
    \vspace{-2px}
\end{figure}%

\vspace{5px}
\noindent
\textbf{MLLMs trail encoder models in terms of classification and abstentnion.} 
When we evaluate the MLLMs across the same taxonomic calibration metrics, they appear to perform slightly worse than to encoder-based models. Among MLLMs, Gemini-2.5 Pro is strongest but still trails BioCLIP significantly (\textrm{IG} $57.7$ vs. $68.9$), indicating current MLLMs underperform specialized encoders for fine-grained, taxonomy-aware classification. Similarly, MLLMs perform worse than to encoder models in terms of abstention tradeoff, with the best MLLM, Gemini-2.5 Pro, lagging $\sim$$3.4$ points behind BioCLIP.

\vspace{5px}
\noindent
\textbf{Range info significantly improves all classification but not abstention.} Realistic bird identification usually makes use of location to restrict the list of possible species under consideration. We wish to gauge the effect of using such a tool on the capability for models to abstain. The effect of this is shown in \cref{fig:range_map_effect}, as well as in detail in \cref{appendix:sec:remaining_figures}. 


We see from \cref{fig:range_map_effect} that using range maps makes the classification problem significantly easier, increasing the average IG across the models $(\mu = 57.2 \rightarrow 88.1)$, while also decreasing the standard deviation substantively $(\sigma = 9.4 \rightarrow 2.1)$. We additionally see this reflected in classification curves of \cref{appendix:sec:remaining_figures}.7, where in particular the genus AUC is near perfect. This is due to the fact that range maps almost always restrict the list species to those within the correct genus.
In contrast, we see a much more mild effect on the abstention tradeoff of the models $(\mu = 45.8 \rightarrow 47.7)$ with an increase in variability $(\sigma = 3.38 \rightarrow 8.02)$. Furthermore, we see that for the MLLMs, abstention tradeoff actually \textit{degrades}. These results support our previous conclusion that these systems need specific improvements for abstention mechanisms, which do not currently exist.




\subsection{Evaluating Abstention Reasoning}


In \cref{fig:brittle_mllm}, we test how robust the \textit{free-form responses} of MLLMs are to small wording changes in instructions. This differs from previous experiments in that, instead of inferring the MLLMs uncertainty using probabilities calculated directly from the language modeling head, we take the MLLM at face value by abstaining when \textit{the text response indicates that a species cannot be predicted}. We follow the same methodology as nlg2choice \cite{lawrence2025you} by generating $15$ equivalent writings for instructions corresponding to three levels of abstention encouragement. This creates a group of $15$ abstention recalls corresponding to each of the three instructions, each indicated by a group of distributions. Each point within the each distribution comprises the abstention rate over the unanswerable dataset given a rewriting of that instruction.

\vspace{.2cm}
\noindent
\textbf{MLLMs have trouble choosing to abstain on their own.} A natural question one might ask is: ``What is the abstention rate of MLLMs in the naive setting?" Namely, one where the user does not explicitly indicate (or know) that an image can be an out-of-schema example. This corresponds to the first set of distributions in \cref{fig:abstention_confusion}, where we find that the default prompt, \texttt{"What species of bird is in this image?"}, has generally low abstention rates. Plainly, if a user does not explicitly tell an MLLM that it can abstain, \textit{it usually will not.}

\vspace{.2cm}
\noindent
\textbf{Abstention rates are extremely brittle.} First, we find that abstention rates vary wildly within a specific group of variations. In particular, Intern3VL has a standard deviation of $\sigma = 10.10$, meaning model abstention can reasonably range from 3\% to 23\%, depending on the minute differences in user writing. Meanwhile, we see that Gemma 3 never abstains, even when exposed to writing variation. Telling models they can abstain generally increases their abstention rates, but can also increase their variability. For instance, Qwen-2.5VL increases abstention rate when telling the model to refuse if unsure $(\mu = 15.29 \rightarrow 66.95)$, but also increases its variability greatly $(\sigma = 7.08 \rightarrow 13.85)$. Strangely, we do not observe this effect by telling the model that the example is potentially unanswerable. Once again, we note that the general exception to this rule is Gemma 3 12B, which has an abstention rate hovering from 1-2\% with tight variance across all instructions. 



\vspace{.2cm}
\noindent
\textbf{MLLMs favor visual-quality explanations and largely ignore missing audio cues.} Across models, the confusion matrices in \cref{fig:abstention_confusion} reveal a strong bias in which abstention reasons are predicted. Models frequently attribute failures to image quality issues (Q), often labeling samples as Q even when the true issue is angle or occlusion (A) or missing vocalization (V). Angle or occlusion problems are also detected to some extent, but far less consistently and with more misclassifications. For example, in Qwen2.5VL, $100\%$ of the detected quality-related failures are correctly labeled as Q, while only $57.6\%$ of angle or occlusion cases are correctly labeled as A, with nearly half of them ($42.4\%$) misclassified as Q. Llama3.2V and InternVL3 show the same pattern of reliably identifying Q but frequently mislabeling A samples as Q. In contrast, virtually no model predicts vocalization as the reason for abstention. Even when the ground truth is V, predictions overwhelmingly fall into Q or A. The systematic under-recognition of V suggests that MLLMs do not treat absent audio cues as a meaningful signal for abstention, likely reflecting the visual bias in training data and limited exposure to audio-dependent failures. 

\begin{table}[t!]
    \centering
    \setlength{\belowcaptionskip}{0pt}
    {\footnotesize
    \captionof{table}{\textbf{Reason-agnostic abstention probability conditioned on ground-truth issue.} Each value reports the probability that a model abstains \emph{for any reason} over all ground truth abstention reasons. Higher values indicate that the model is more likely to abstain when that type of issue is present, regardless of which abstention label it chooses.}
    \vspace{-15px}
    \begin{tabular}{lccc}
    & \multicolumn{3}{c}{\textbf{Ground-Truth Issue}} \\
    \cmidrule(lr){2-4}
    \textbf{Model} & \textbf{quality} & \textbf{angle/occlusion} & \textbf{vocalization} \\
    \midrule
    Qwen2.5-VL-7B      & 0.158 & 0.144 & 0.098 \\
    Llama-3.2V-11B     & 0.086 & 0.080 & 0.077 \\
    InternVL3-8B       & \textbf{0.279} & \textbf{0.291} & \textbf{0.278} \\
    Gemma 3 12B        & 0.041 & 0.052 & 0.044 \\
    \bottomrule
    \end{tabular}
    \label{table:abstention_confusion_recall}
    }
\vspace{-15px}
\end{table}




\vspace{-2px}
\section{Conclusion}
We introduced RealBirdID, a benchmark for bird species classification with abstention. We assembled a dataset of expert-vetted iNaturalist images where the species cannot be determined from the image alone, either for a priori reasons such as missing vocalizations, range, or temporal information, or for visual reasons such as low image quality, occlusion, or extreme viewpoint. Our experiments show that current encoders and MLLMs fail to abstain reliably on these examples, and even when they abstain their stated reasons are often incorrect. We hope this benchmark will support the development of fine-grained recognition systems that are both able to abstain when appropriate and ultimately more robust when deployed in real world applications.

\section*{Acknowledgements}
The project was supported in part by National Science Foundation award \#2329927.
This research was done using services provided by the OSG Consortium \cite{ospool1, ospool2, ospool3, ospool4}, which is supported by the National Science Foundation awards \#2030508 and \#2323298.




{
    \small
    \bibliographystyle{ieeenat_fullname}
    \bibliography{main}
}

%
\clearpage
\setcounter{page}{1}
\maketitlesupplementary
\appendix

\section{Different Abstention Thresholding Criteria}\label{appendix:sec:metric_details}

As mentioned in the encoder abstention metrics section, the choice of max probability to create a threshold for abstention tradeoff is not obvious. In this section, we experiment with two different methods for creating a decision threshold: (1) entropy, (2) max probability, and (3) probability difference.

\begin{figure*}[ht!]
    \centering
    \captionsetup{type=figure}
    \setlength{\belowcaptionskip}{0pt}
    \includegraphics[width=\linewidth]{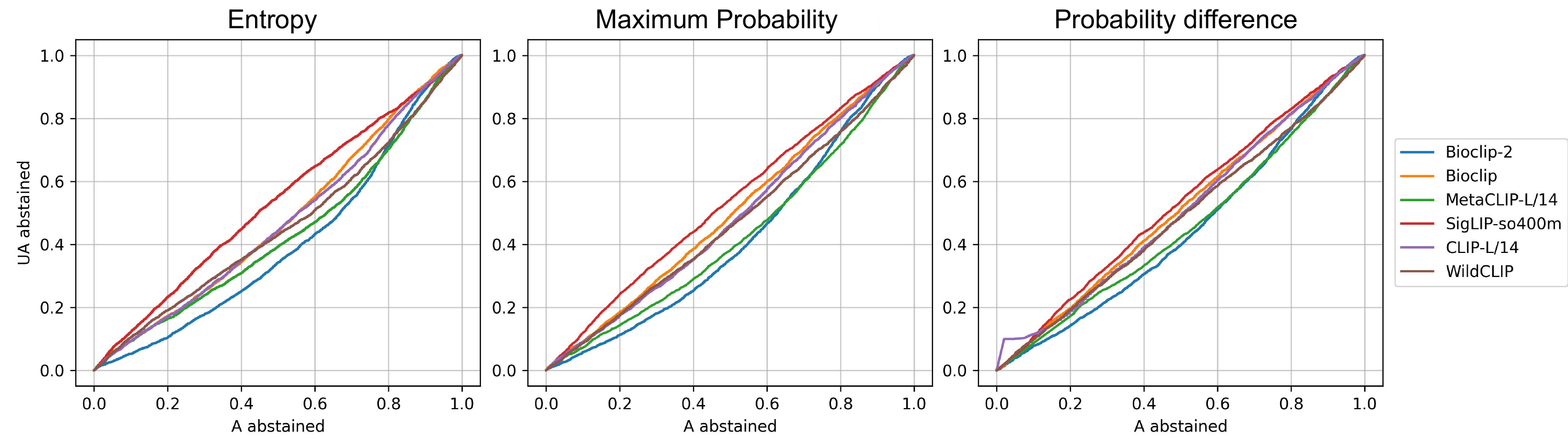}
    \vspace{0pt}
    \captionof{figure}{\textbf{Abstention tradeoff curves for different abstention criteria.}}
    \label{fig:abstention_tradeoff_ablation_criteria}
\end{figure*}

\begin{figure*}[h!]
    \centering
    \captionsetup{type=figure}
    \setlength{\belowcaptionskip}{0pt}
    \includegraphics[width=\linewidth]{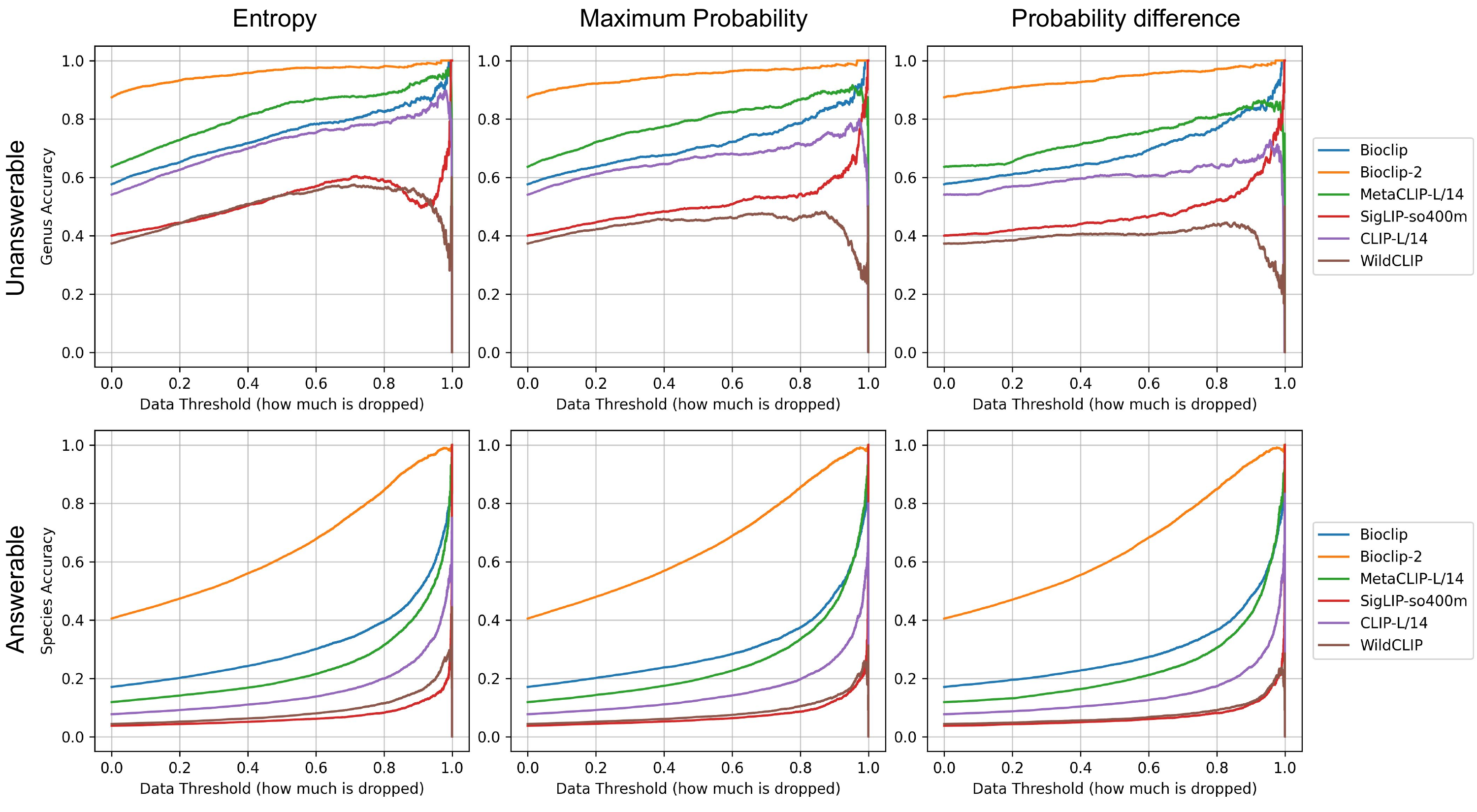}
    \vspace{0pt}
    \captionof{figure}{\textbf{Accuracy vs thresholding for different abstention criteria.}}
    \label{fig:classification_tradeoff_ablation_criteria}
\end{figure*}

\paragraph{The difference between criteria is small.} In \cref{fig:abstention_tradeoff_ablation_criteria} we show the abstention tradeoff for various models using the three criteria. We find that in all cases, models hover around random prediction with the best results being achieved by probability difference. Similarly, in \cref{fig:classification_tradeoff_ablation_criteria} we see that varying the criteria for these models does not significantly affect classification performance, with the best two criteria being max probability and probability difference. 

\begin{figure*}[ht!]
    \centering
    \captionsetup{type=figure}
    \setlength{\belowcaptionskip}{0pt}
    \includegraphics[width=\linewidth]{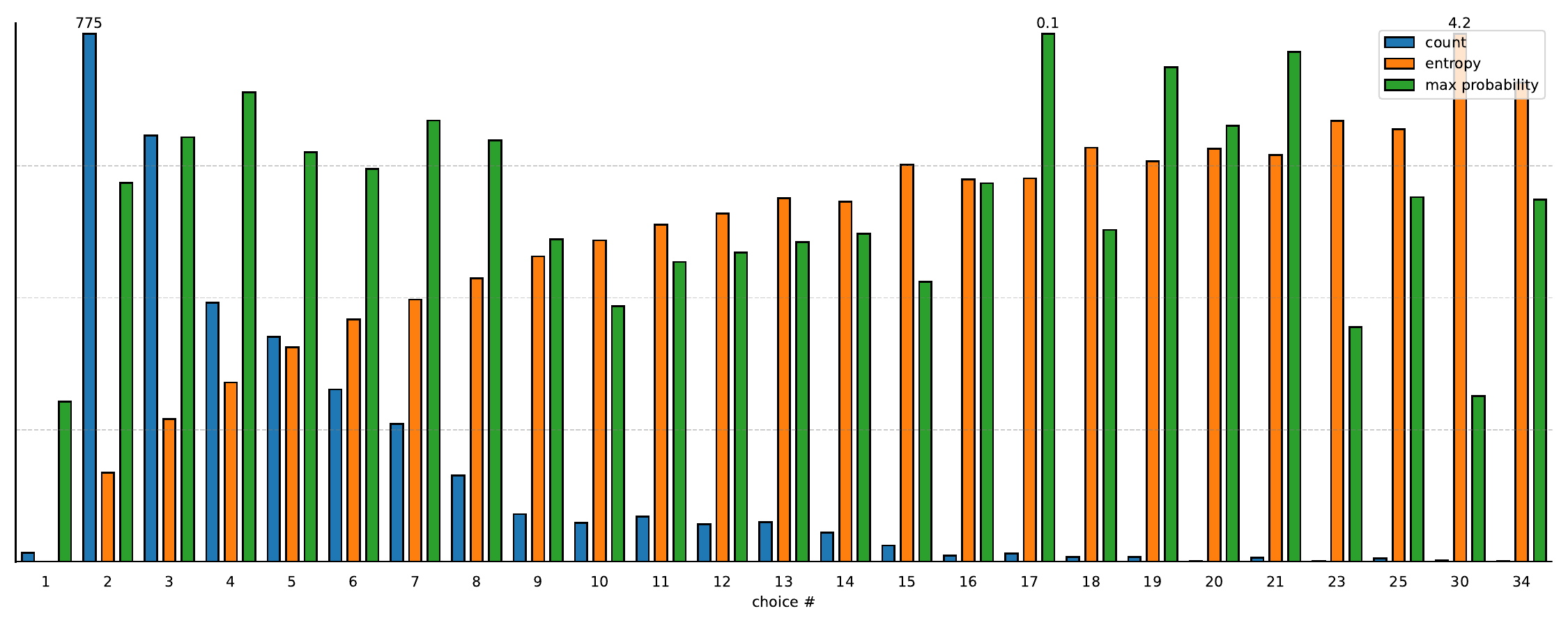}
    \includegraphics[width=\linewidth]{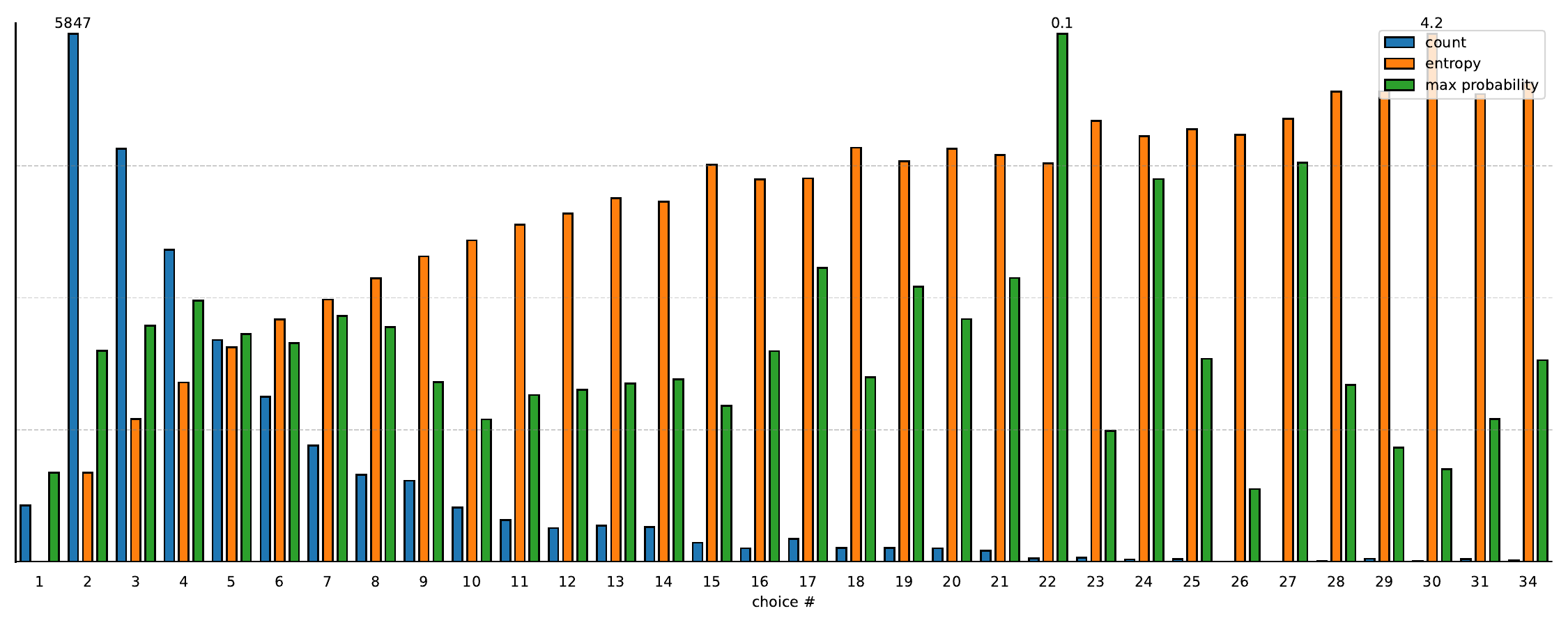}
    \vspace{0pt}
    \captionof{figure}{\textbf{Frequency, probability entropy, and max probability for various choice counts in the Range Info Species Subset data.} For the Range Info data, the examples are split up by the amount of choices that the model must choose between. For each choice count, the frequency, entropy, and max probability are shown. Bars are normalized by series. (\textbf{Top}) depicts the answerable set, whereas (\textbf{bottom}) depicts the unanswerable set. Both are run with Qwen-2.5VL-7B.}
    \label{fig:max_idea}
\end{figure*}

\paragraph{Max probability is less correlated with choice count.} A large portion of our experiments involve subsetting the possible list of species for a given observation by SINR. Naturally, this introduces a discrepancy in the amount of choices depending on the observation. We hypothesize that this caused an effect on the average entropy for each choice count, depicted in \cref{fig:max_idea}. There, we see that entropy increases as a steady function of choice count. However, we see that max probability does not have a clear correlation with choice count. This means that using max probability as a criteria less susceptible to simply predicting images with higher species counts as confusing in the abstention tradeoff. 

\section{Additional Experiment Details}\label{appendix:sec:additional_experiments}

\paragraph{Querying Proprietary APIs.} For the proprietary MLLMs GPT-5 and Gemini-2.5 Pro, we use the single-turn API calls using the same prompt \texttt{"What is the species of this bird?"} and the input image. No system prompts, auxiliary instructions, or tool calls were used. Images were transmitted as base64-encoded JPEGs according to the providers’ multimodal specifications. We set the reasoning-effort parameter to its minimal value (\texttt{minimal} for both models) and kept all other generation parameters at their documented defaults, including the default sampling temperature and no maximum output token limit. In this configuration, GPT-5 typically produced responses with negligible explicit reasoning tokens, whereas Gemini-2.5 Pro often emitted a short chain-of-thought before the final answer (87 tokens on average), resulting in a modest but consistent reasoning-token overhead. Across the answerable (31{,}885) and unanswerable (3{,}253) subsets, we issued 35{,}138 multimodal API calls per model.

\paragraph{BioCLIP-2 leads the pack in terms of classification performance.} We find the best performing encoder to be BioCLIP-2, having the highest species classification abilities $(AUC = \textbf{.567})$, genus classification abilities $(AUC = \textbf{.934})$, and information gain $(AUC = \textbf{.845})$. In particular, we find the out-of-the-box accuracies on the species ($3561$-way) and genus ($248$-way) levels to be \textbf{41\%} and \textbf{76\%} respectively, very strong performance for such a large multi-way task.

\paragraph{Training data leakage.} To assess whether these results could be attributed to training data leakage, we conducted a comprehensive overlap analysis between BioCLIP-2's training corpus (TreeOfLife-200M) and our test sets. Each test image is associated with an iNaturalist observation ID, observation URL, and photo URL, while each TOL image contains a single \texttt{source\textunderscore url} field. We canonicalized all identifiers on both sides, mapping each test-set identifier and each TOL \texttt{source\textunderscore url} to a normalized photo ID, observation ID, or URL, and compared these canonical keys for exact and near-exact matches. Here, ``near-exact'' refers to different URL forms of the same underlying media (e.g., different resolutions or hostnames that map to the same iNaturalist photo or observation ID). Across the full $213.9$M TOL images, we identified overlaps for $56.5\%$ of samples in the answerable test set ($18{,}007/31{,}885$ images) and $1.1\%$ in the unanswerable set ($36/3{,}253$ images). This substantial overlap suggests possible training data leakage and may partially explain BioCLIP-2's high performance. These findings highlight the need for future evaluations on fully de-duplicated benchmarks to more rigorously assess out-of-distribution generalization.




\section{Additional Details}
\paragraph{Ethical and Licensing Considerations.} This release is intended for validation and analysis only. We used whatever media licenses were returned by the API; in the public release, we will (i) filter to permitted licenses (e.g., CC-BY/CC-BY-NC) and (ii) include clear provenance to original observations, respecting iNaturalist’s terms and any location obscuration for sensitive taxa. Because we do not provide training splits, we also avoid any leakage between unanswerable and answerable resources by not reusing the exact observation images across sections.

\paragraph{Limitations.} Firstly, the unanswerable dataset is inherently imbalanced: certain genera are overrepresented due to uneven observation rates, and some species pairs are more prone to visual ambiguity than others. This imbalance may influence both model behavior and evaluation metrics. Second, the labeling of unanswerable examples depends on expert judgment. For example, experts may disagree on what constitutes an "obstructed view" or whether a particular image lacks sufficient evidence for identification.

Future work could extend this framework to multimodal settings, incorporating optional modalities such as sound recordings, temporal context, and multiple observations. Another direction is improving the abstention calibration, particularly for multimodal MLLMs, which tend to overcommit despite uncertainty.

Although this work aims to promote responsible deployment through abstention-aware modeling, potential negative societal impacts should be considered. Miscalibrated abstention or overconfident misclassification may undermine public trust in these tools, especially when integrated into citizen-science platforms. Overreliance on model outputs could also discourage human expertise or misinform conservation decisions if abstention signals are misunderstood. To mitigate these risks, future iterations of the benchmark and accompanying systems should emphasize transparency, interpretability, and human-in-the-loop evaluation.

\newpage

\section{Organization of Remaining Figures.}\label{appendix:sec:remaining_figures}

\begin{figure*}[ht!]
    \centering
    \captionsetup{type=figure}
    \setlength{\belowcaptionskip}{0pt}
    \includegraphics[width=\linewidth]{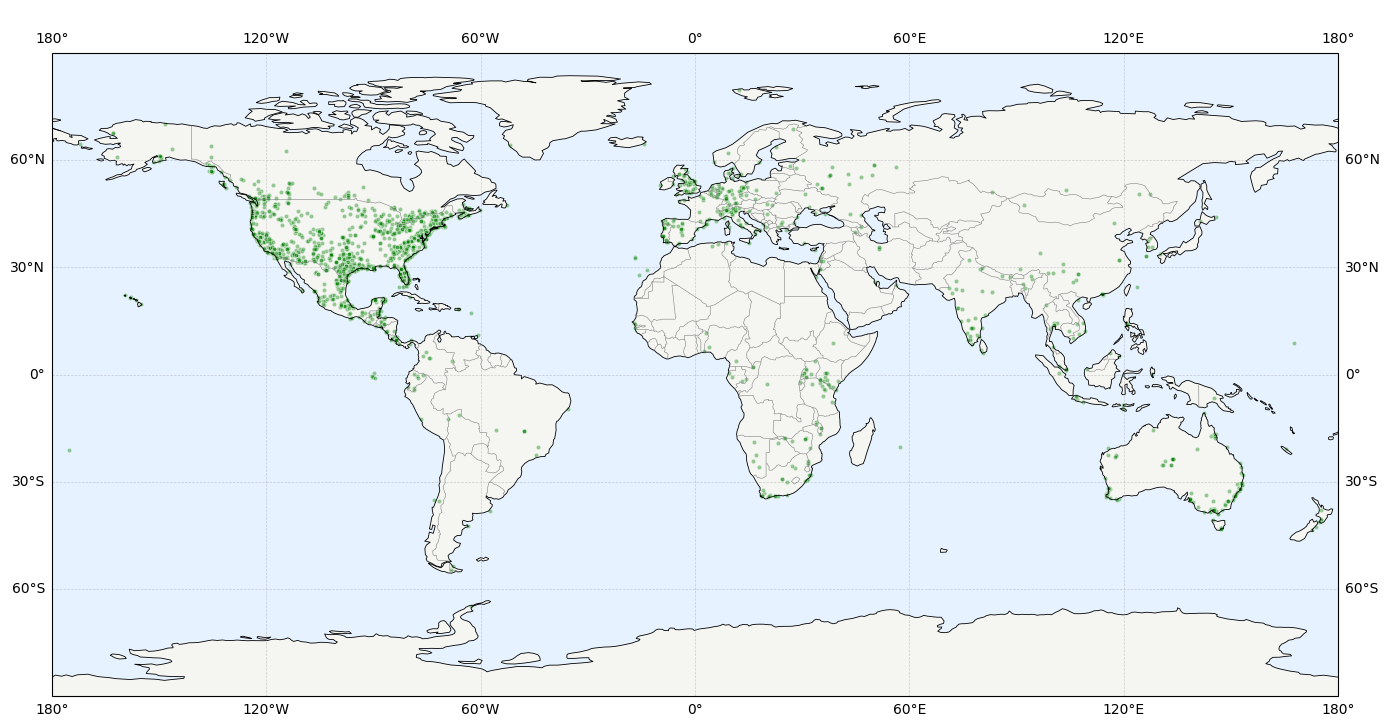}
    \includegraphics[width=\linewidth]{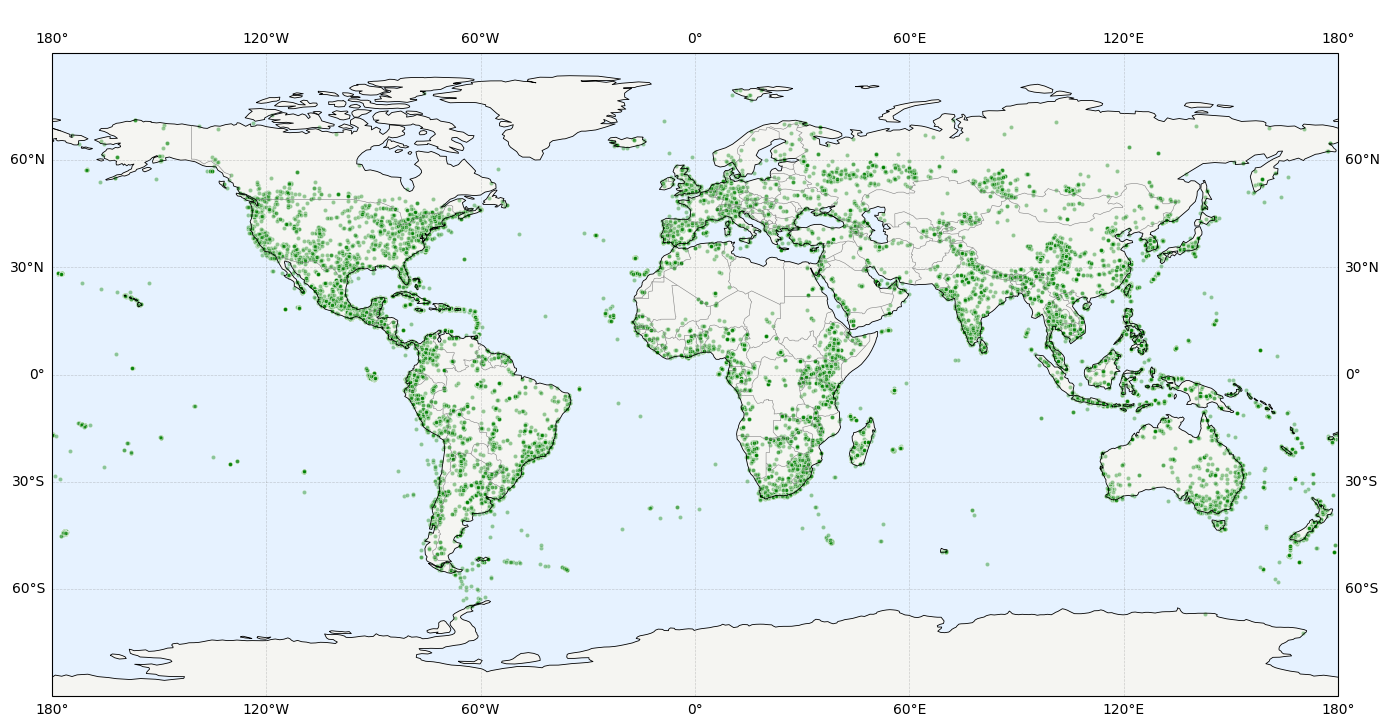}
    \vspace{0pt}
    \captionof{figure}{\textbf{Location info of the Unanswerable (top) and Answerable (bottom) sets.}}
    \label{fig:coordinate_data}
\end{figure*}

\begin{figure*}[ht!]
    \centering
    \captionsetup{type=figure}
    \setlength{\belowcaptionskip}{0pt}
    \includegraphics[width=\linewidth]{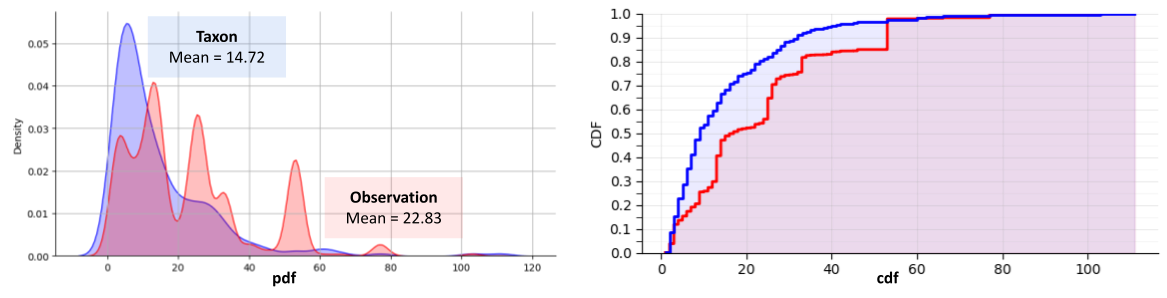}
    \vspace{0pt}
    \captionof{figure}{\textbf{Distribution of species per observation and per taxon.}}
    \label{fig:species_per_obs_taxon}
\end{figure*}

\begin{figure*}[ht!]
    \centering
    \captionsetup{type=figure}
    \setlength{\belowcaptionskip}{0pt}
    \includegraphics[width=\linewidth]{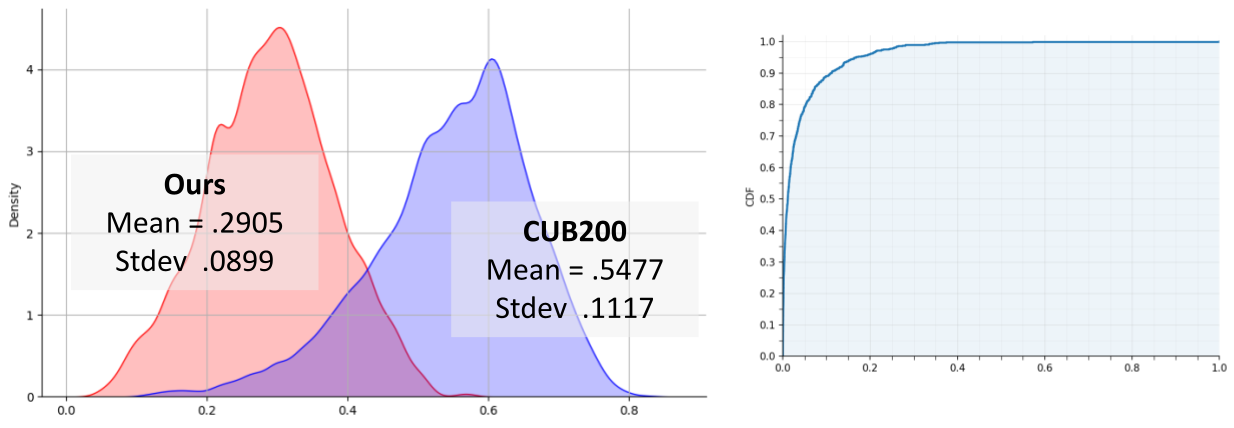}
    \vspace{0pt}
    \captionof{figure}{\textbf{MANIQA distributions of RealBirdID (ours) vs. CUB200.} On average, we find that RealBirdID has lower MANIQA scores.}
    \label{fig:maniqa_realbirdid_cub200}
\end{figure*}

\begin{figure*}[ht!]
    \centering
    \captionsetup{type=figure}
    \setlength{\belowcaptionskip}{0pt}
    \includegraphics[width=\linewidth]{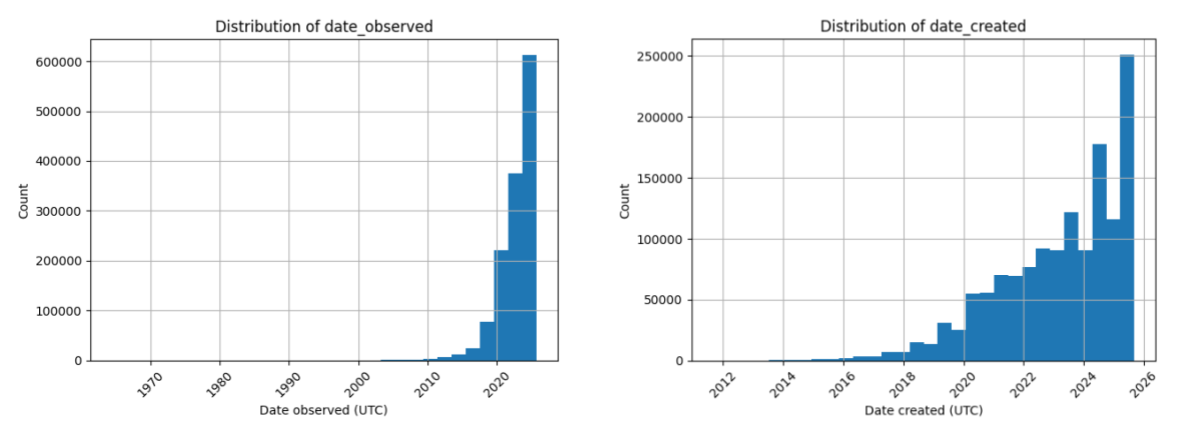}
    \vspace{0pt}
    \captionof{figure}{\textbf{Observation date distribution by observed and created timestamps.}}
    \label{fig:date_distribution_realbirdid}
\end{figure*}

\begin{figure*}[ht!]
    \centering

    \begin{minipage}{0.48\linewidth}
        \centering
        \includegraphics[width=\linewidth]{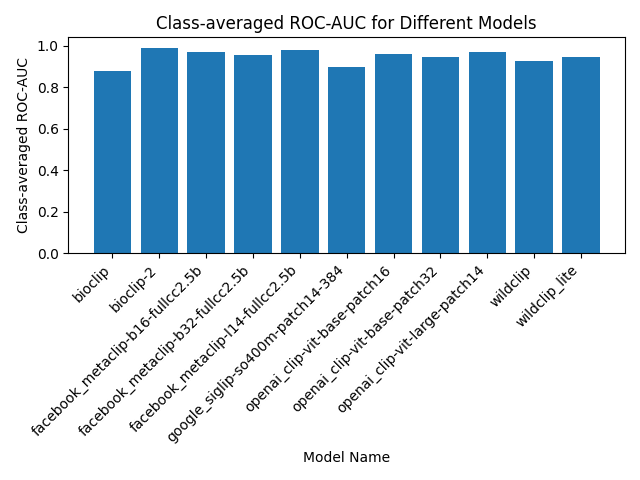}
        \caption*{Binary Classification Results on the Answerable Subset (Left)}
    \end{minipage}
    \hfill
    \begin{minipage}{0.48\linewidth}
        \centering
        \includegraphics[width=\linewidth]{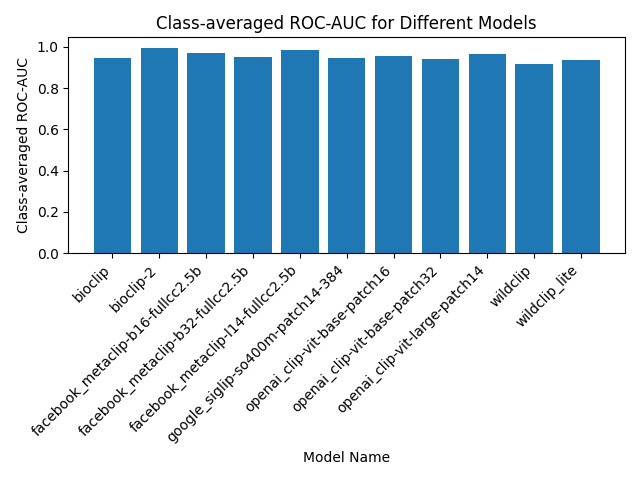}
        \caption*{Binary Classification Results on the Unanswerable Subset (Right)}
    \end{minipage}
    \caption{\textbf{Classification results using class-averaging on the answerable and unanswerable sets.}}
    \label{fig:classification_class_averaged}
\end{figure*}

\paragraph{1. Spacial distribution of observations seen within the answerable and unanswerable data.} \cref{fig:coordinate_data}
\paragraph{2. Distribution of species per observation and per taxon.} \cref{fig:species_per_obs_taxon}
\paragraph{3. MANIQA Distribution of RealBirdID vs. CUB200} \cref{fig:maniqa_realbirdid_cub200}
\paragraph{4. RealBirdID iNaturalist observation dates.} \cref{fig:date_distribution_realbirdid}
\paragraph{5. Table of most common genera occuring in the unanswerable data.} \cref{table:most_common_genera}
\paragraph{6. Classification results using class-averaging on the answerable and unanswerable sets.} \cref{fig:classification_class_averaged}
\paragraph{7. Detailed classification curve tradeoffs and compiled range map information.} \cref{fig:classification_curves_q1}, \cref{fig:abstention_classification_curves}, \cref{fig:encoder_classification_abstention_performance}
\paragraph{8. 100 examples randomly chosen from the answerable set.} \cref{fig:answerable_example_grid}
\paragraph{9. 30 unanswerable examples which have abstention reason of ``angle/occlusion."} \cref{fig:angle_ua_example_grid}
\paragraph{10. 30 unanswerable examples which have abstention reason of ``vocalization."} \cref{fig:vocalization_ua_example_grid}
\paragraph{11. 30 unanswerable examples which have abstention reason of ``angle/occlusion."} \cref{fig:quality_ua_example_grid}



\begin{table*}[h]
\setlength{\belowcaptionskip}{-20pt}
\setlength{\tabcolsep}{15pt}
\begin{center}
\begin{tabular}{p{7.8cm}rrrr}
\hline
\textbf{genus} & \textbf{count} & \textbf{freq} & \textbf{cdf} & \textbf{species} \\
\hline
Crows and Ravens & 291 & 0.1261 & 0.1261 & 53 \\
Large White-headed Gulls & 198 & 0.0858 & 0.2119 & 25 \\
Kingbirds & 179 & 0.0776 & 0.2894 & 13 \\
Empidonax Flycatchers & 132 & 0.0572 & 0.3466 & 14 \\
Mallards, Pintails, and Allies & 125 & 0.0542 & 0.4008 & 33 \\
Dryobates Woodpeckers & 116 & 0.0503 & 0.4510 & 26 \\
Rufous, Allen's, and Allied Hummingbirds & 53 & 0.0230 & 0.4740 & 9 \\
Buteos & 52 & 0.0225 & 0.4965 & 27 \\
Dowitchers & 47 & 0.0204 & 0.5169 & 3 \\
Calidris Sandpipers & 39 & 0.0169 & 0.5338 & 24 \\
Scaups, Pochards, and Allies & 39 & 0.0169 & 0.5507 & 12 \\
True Swans & 37 & 0.0160 & 0.5667 & 9 \\
Leaf Warblers & 35 & 0.0152 & 0.5819 & 77 \\
Yellow-breasted Meadowlarks & 32 & 0.0139 & 0.5958 & 3 \\
Shanks, Tattlers, and Allies & 27 & 0.0117 & 0.6075 & 13 \\
Chickadees and Allies & 27 & 0.0117 & 0.6192 & 15 \\
Ruby-throated and Black-chinned Hummingbirds & 25 & 0.0108 & 0.6300 & 2 \\
Plegadis Ibises & 24 & 0.0104 & 0.6404 & 3 \\
Typical Falcons & 23 & 0.0100 & 0.6503 & 40 \\
American Cormorants & 20 & 0.0087 & 0.6590 & 3 \\
Brown Thrushes and Nightingale-Thrushes & 20 & 0.0087 & 0.6677 & 13 \\
Great Herons and Egrets & 19 & 0.0082 & 0.6759 & 17 \\
Western and Clark's Grebes & 19 & 0.0082 & 0.6841 & 2 \\
Setophaga Warblers & 19 & 0.0082 & 0.6924 & 34 \\
Yellow-tailed and White-tailed Black Cockatoos & 18 & 0.0078 & 0.7002 & 3 \\
Vireos & 17 & 0.0074 & 0.7075 & 33 \\
True Sparrows & 15 & 0.0065 & 0.7140 & 28 \\
\multicolumn{5}{c}{\dots} \\
Trillers and Allies & 1 & 0.0004 & 0.9991 & 20 \\
Typical White-eyes & 1 & 0.0004 & 0.9996 & 111 \\
Locustellid Bush Warblers and Allies & 1 & 0.0004 & 1.0000 & 23 \\
\hline
\end{tabular}
\caption{\textbf{Most common generas occuring in the unanswerable data.}}
\label{table:most_common_genera}
\end{center}
\end{table*}

\newpage
\begin{figure}[t!]
    \centering
    \captionsetup{type=figure}
    \setlength{\belowcaptionskip}{-15pt}
    \includegraphics[width=\linewidth]{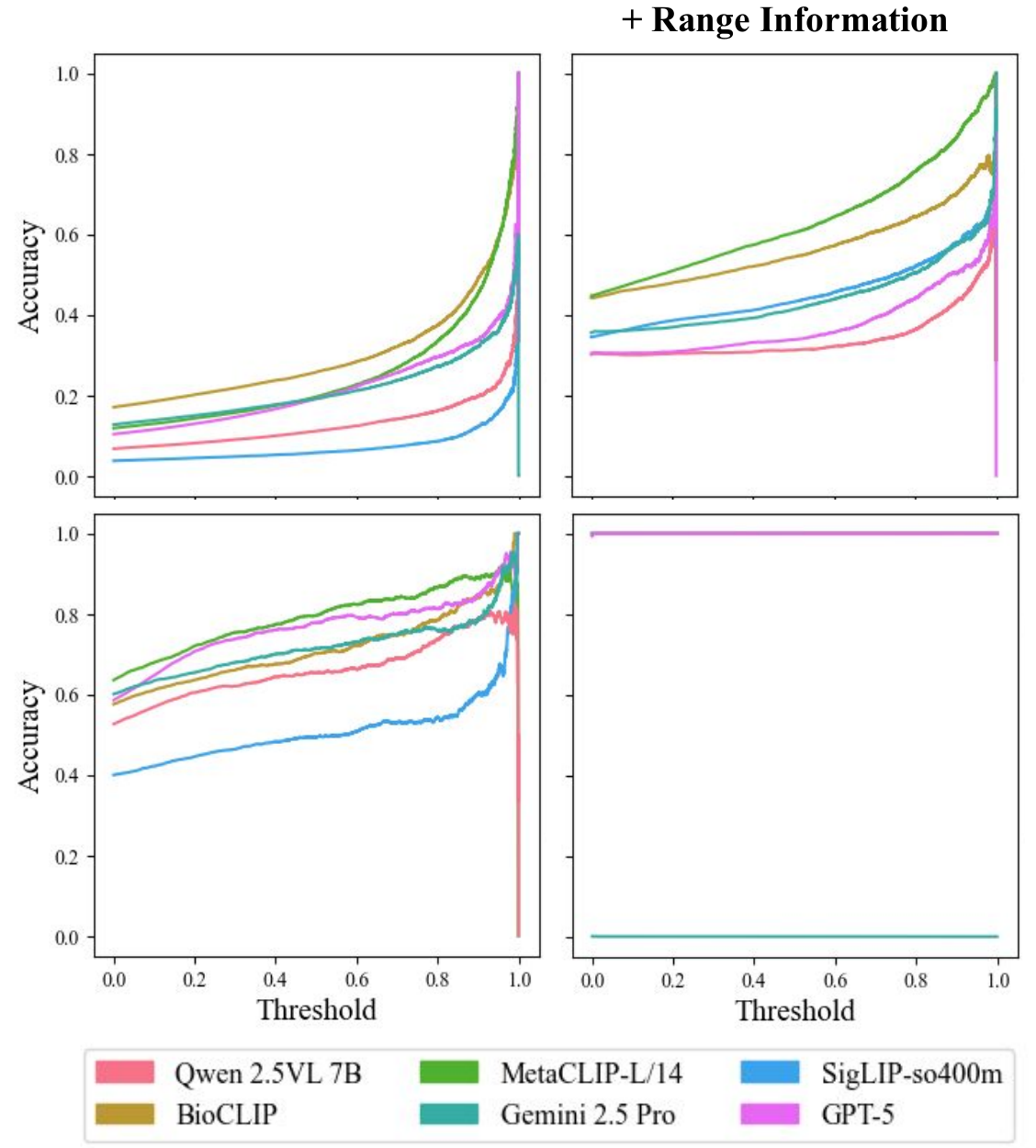}
    \vspace{-10pt}
    \captionof{figure}{\textbf{Classification performance of CLIP-based models and popular MLLMs on the species level of answerables (A) and the genus level of unanswerables (UA).} For various CLIP encoders, accuracies at percentile-based max probablities are plotted when sweeping over percent of data thresholded. For the answerable set the species label is used to compute accuracy \textbf{(top)} whereas for the unanswerable set, the genus label is used \textbf{(bottom)}. On the \textbf{(right)} we observe the effect of using \textit{species range maps} to constrict the choice set. Note that the genus accuracy for encoders is \emph{not an error}: subsetting species list using SINR location info achieves perfect genus performance. }
    \label{fig:classification_curves_q1}
\end{figure}
\begin{figure}[t!]
    \centering
    \captionsetup{type=figure}
    \setlength{\belowcaptionskip}{-10pt}
    \includegraphics[width=\linewidth]{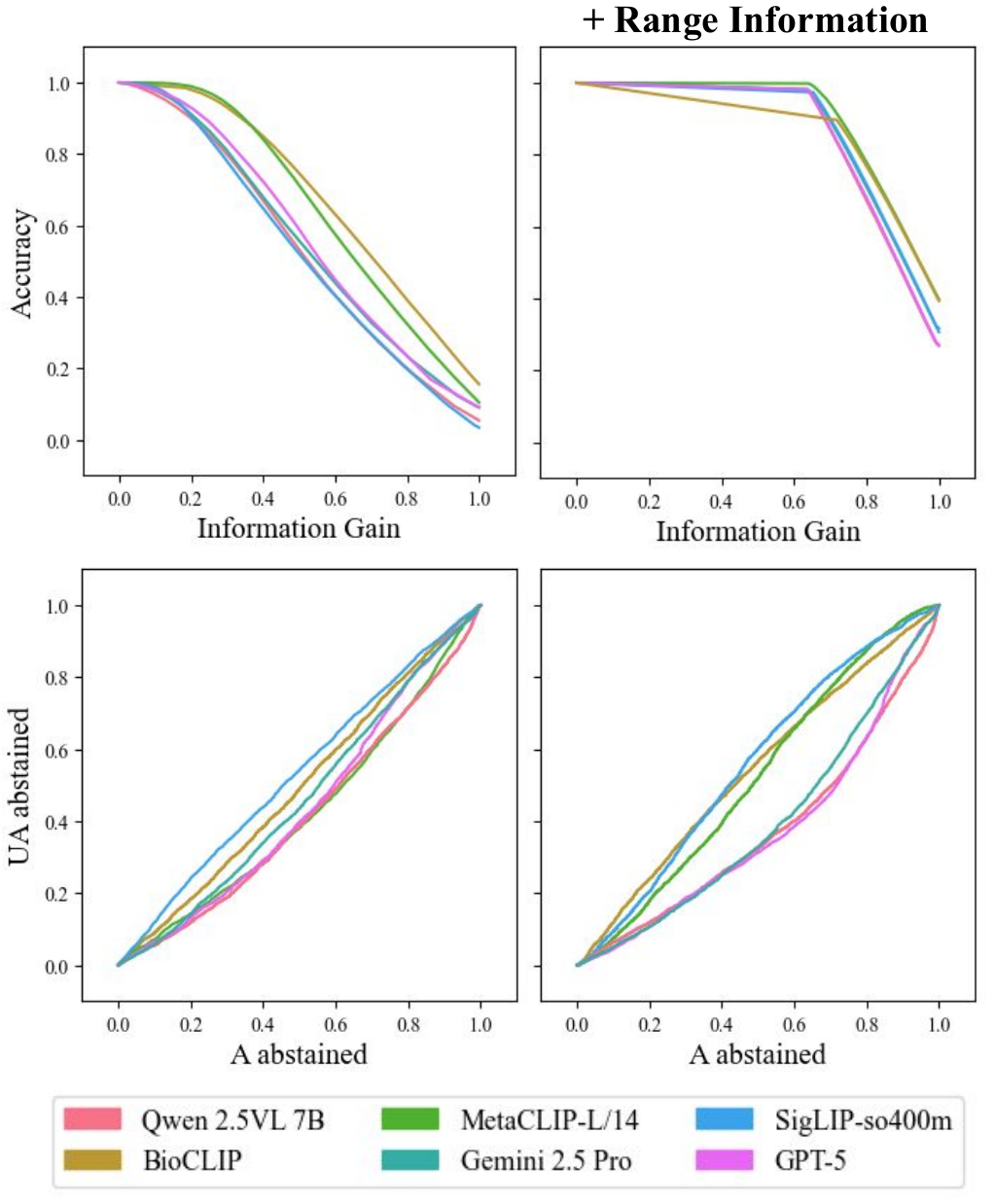}
    \vspace{-10pt}
    \captionof{figure}{\textbf{Abstention calibration and entropy–threshold selective classification.} For each model we sweep an entropy threshold on the flat species-level softmax and plot the fraction of unanswerable (UA) examples abstained on against the fraction of answerable (A) examples abstained \textbf{(bottom)}. We combine the UA / A performance into a unified classification metric using Information Gain vs. Accuracy as proposed by DARTS (\textbf{top}).}
    \label{fig:abstention_classification_curves}
\end{figure}

\begin{figure*}[ht!]
    \centering
    \captionsetup{type=figure}
    \setlength{\belowcaptionskip}{-10pt}
    \includegraphics[width=\linewidth]{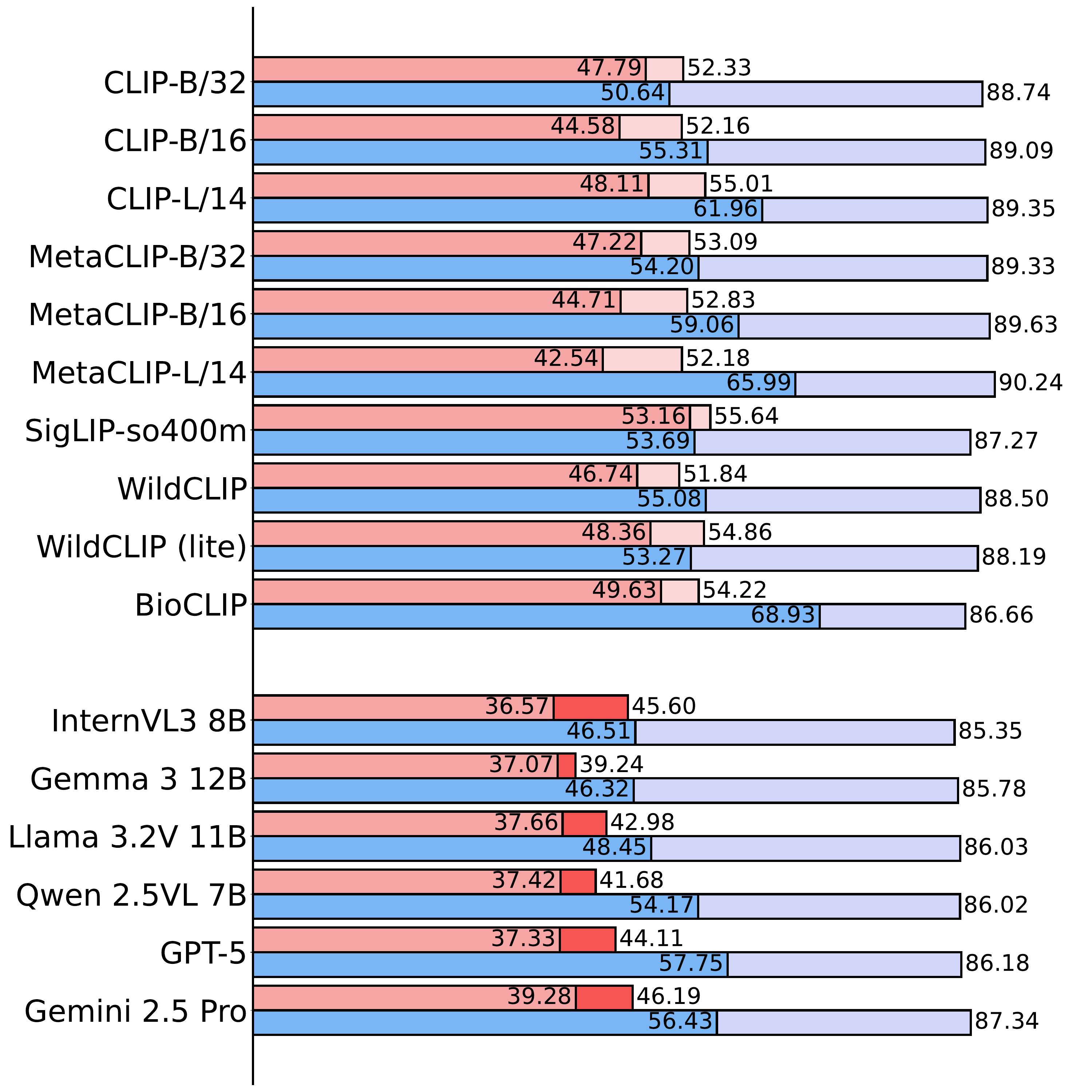}
    \captionof{figure}{\textbf{Summary of classification and abstention performance on RealBirdID.} Blue bars show classification performance (IG), a metric which mixes species and genus-level accuracy over coverage on answerable (A) examples, while red bars show abstention capability, measured as AUROC for separating answerable (A) from unanswerable (UA) instances (higher is better for both). The lighter bars correspond to increases in performance from using \textit{species range maps} whereas the darker bars indicate performance decreases. Notably, abstention tradeoff performance decreases for MLLMs.
    }
    \label{fig:encoder_classification_abstention_performance}
\end{figure*}


\newpage
\begin{figure*}[ht!]
    \centering
    \captionsetup{type=figure}
    \setlength{\belowcaptionskip}{0pt}
    \includegraphics[width=\linewidth]{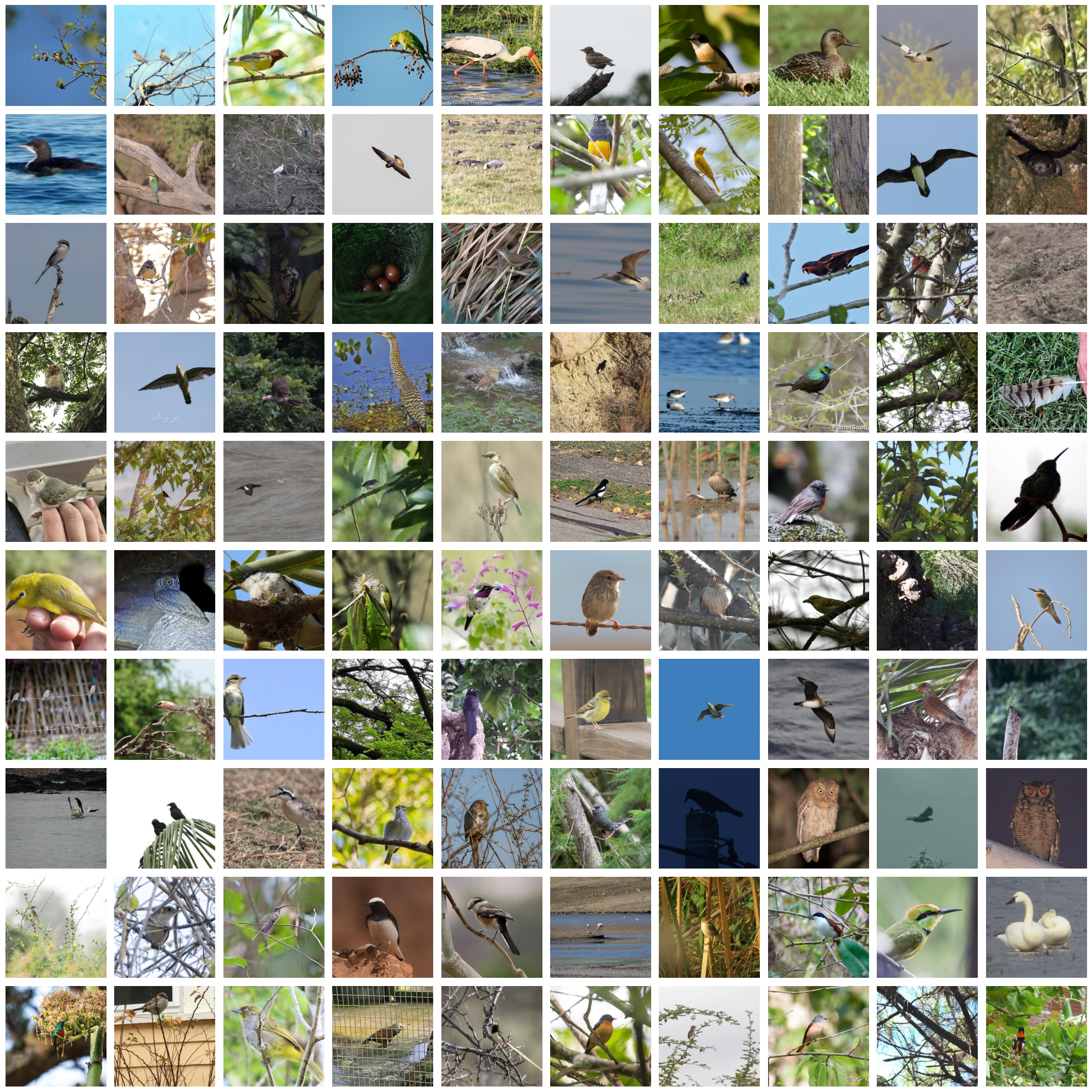}
    \vspace{0pt}
    \captionof{figure}{\textbf{100 examples of images from the answerable set.} The answerable set is sampled from Research Grade iNaturalist images to fill out species corresponding to sampled Unanswerable data.}
    \label{fig:answerable_example_grid}
\end{figure*}

\newpage
\begin{figure*}[ht!]
    \centering
    \captionsetup{type=figure}
    \setlength{\belowcaptionskip}{0pt}
    \includegraphics[width=\linewidth]{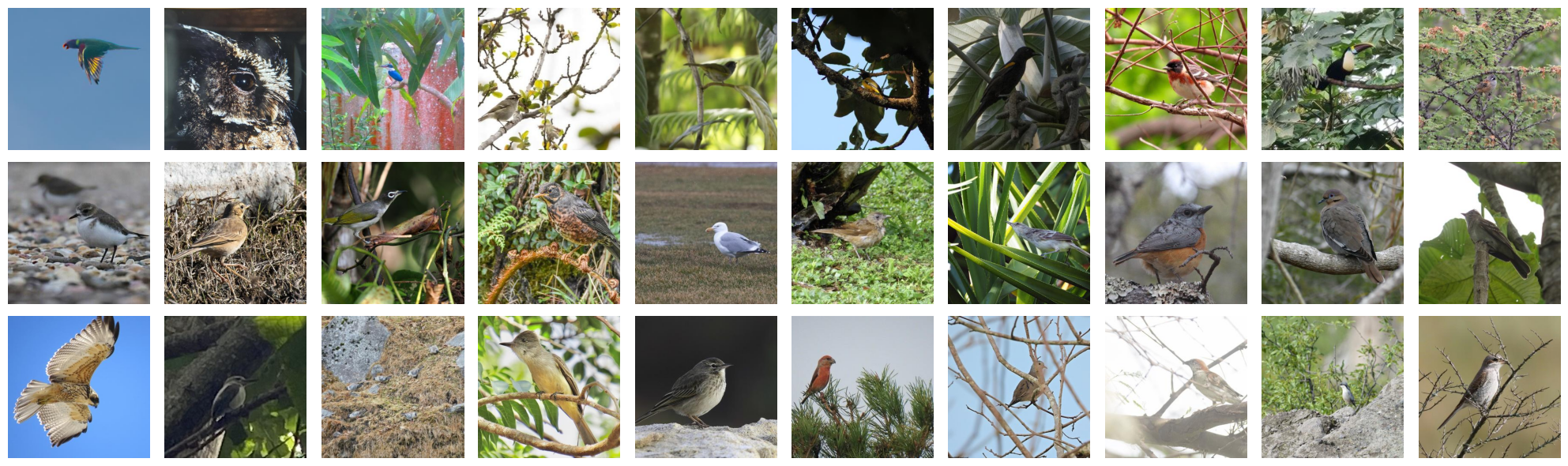}
    \vspace{0pt}
    \captionof{figure}{\textbf{30 images with \textit{angle/occlusion} abstention reasons from the unanswerable set.}}
    \label{fig:angle_ua_example_grid}
\end{figure*}
\begin{figure*}[ht!]
    \centering
    \captionsetup{type=figure}
    \setlength{\belowcaptionskip}{0pt}
    \includegraphics[width=\linewidth]{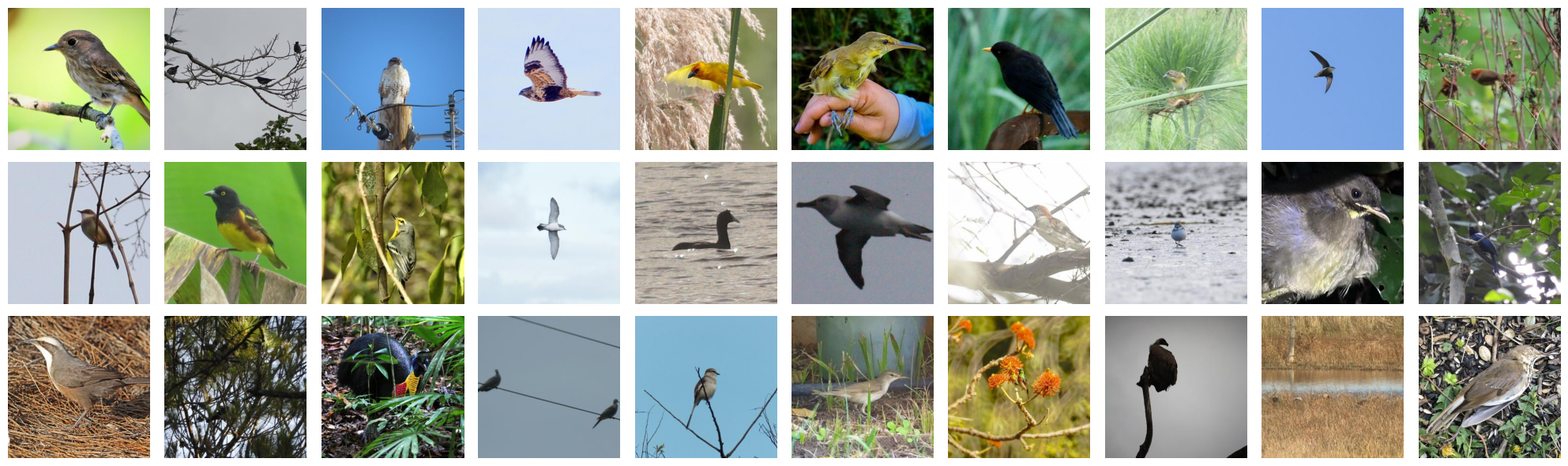}
    \vspace{0pt}
    \captionof{figure}{\textbf{30 images with \textit{vocalization} abstention reasons from the unanswerable set.}}    \label{fig:vocalization_ua_example_grid}
\end{figure*}
\begin{figure*}[ht!]
    \centering
    \captionsetup{type=figure}
    \setlength{\belowcaptionskip}{0pt}
    \includegraphics[width=\linewidth]{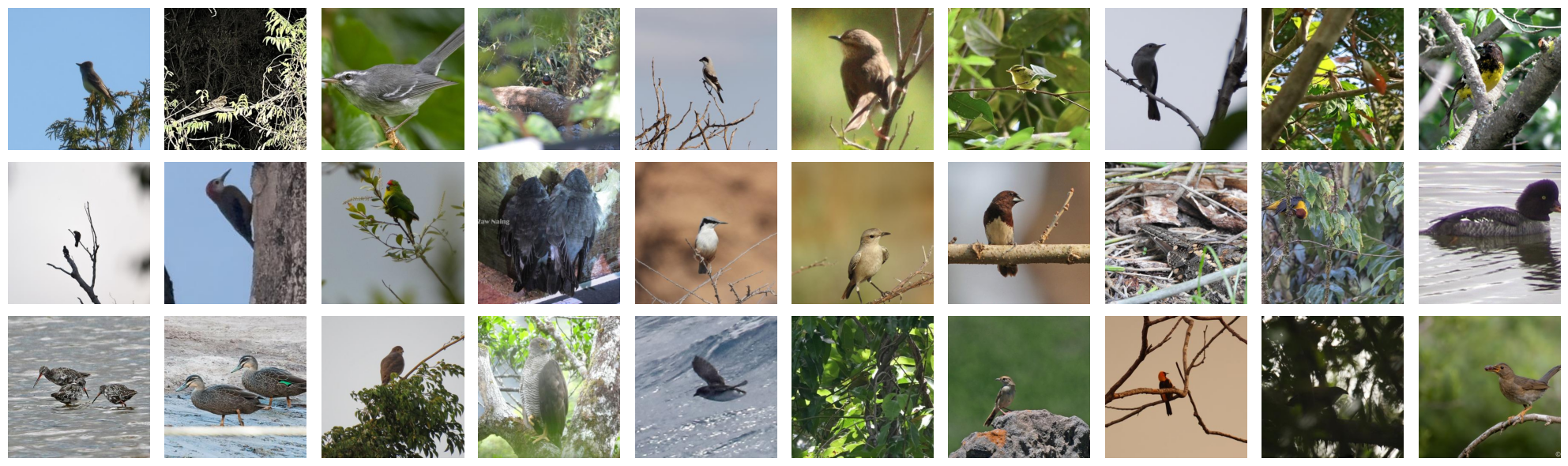}
    \vspace{0pt}
    \captionof{figure}{\textbf{30 images with \textit{image quality} abstention reasons from the unanswerable set.}}
    \label{fig:quality_ua_example_grid}
\end{figure*}

\end{document}